\documentclass[a4paper,10pt]{article}
\usepackage{graphicx}
\usepackage{xcolor}

\title{Assessment of the Local Tchebichef Moments Method for Texture Classification by Fine Tuning Extraction Parameters}
\author{Andre Barczak \textsuperscript{1,*} \and Napoleon Reyes \textsuperscript{1} \and Teo Susnjak \textsuperscript{1}}

\begin{document}

\maketitle

\textbf{1-Massey University, Auckland, New Zealand}

\textbf{* a.l.barczak@massey.ac.nz}

ACM-class: I.4.9; I.5.4

\begin{abstract}
In this paper we use machine learning to study the application of Local Tchebichef Moments (LTM) to the problem of texture classification.
The original LTM method was proposed by Mukundan~\cite{Mukundan2014}.

The LTM method can be used for texture analysis in many different ways, either using the moment values directly, or more simply creating a relationship between the moment values of different orders, producing a histogram similar to those of Local Binary Pattern (LBP) based methods.
%To the best of our knowledge, this is the first time that this method was systematically used for performance assessment of texture classification.
The original method was not fully tested with large datasets, and there are several parameters that should be characterised for performance.
Among these parameters are the kernel size, the moment orders and the weights for each moment.

We implemented the LTM method in a flexible way in order to allow for the modification of the parameters that can affect its performance.
Using four subsets from the Outex dataset (a popular benchmark for texture analysis), we used Random Forests to create models and to classify texture images, recording the standard metrics for each classifier.
We repeated the process using several variations of the LBP method for comparison.
This allowed us to find the best combination of orders and weights for the LTM method for texture classification.
\end{abstract}
\section{Introduction}
Texture classification is a classic problem in machine vision and pattern recognition.
A large number of methods exist, and in the last few years many new feature extraction methods were developed targeting fast extraction and improved classification accuracy.
Two recent reviews on feature extraction methods for texture analysis were carried out by Humeau-Heutier \cite{Heurtier2019} and Liu et al. \cite{Liu2019}.
%In this work the focus is on a particular type of feature extraction for texture classification, which involves the computation of values for each pixel of the image using a certain function.

A very popular method for texture feature extraction is called Local Binary Pattern (LBP). This method uses the values of neighbourhood pixels to create a binary code for the central pixel.
An image can be created using such a method, e.g., if the 8 neighbour pixels are used, it creates a code with $2^8$ values, which fits into a grayscale image.
The histogram of the LBP image can be used directly as a feature space for classification purposes.

%TODO two recent papers carried out a broad and complete review of texture techniques and ... \cite{Liu2019} \cite{Heurtier2019}

Liu et al. also did an evaluation of LBP compared to Deep Texture Descriptors \cite{LiuEtAl2016}. 
One of the findings is that CNN methods are very accurate but their computational complexity and memory requirements are costly.
Often engineered features are able to achieve similar accuracy, but in real-time.

Silva, Bouwmans and Frelicot \cite{SilvaEtAl2015} carried out an extensive experimentation with LBP and its variants.
They proposed and experimented with a new LBP variant called extended center-symmtric LBP (XCS-LBP), which achieved very hight accuracy for background modelling problems.

Mukundan has proposed several feature extraction methods based on the Tchebichef polynomials.
In \cite{MukundanEt01} a new Tchebichef moment method was proposed.
A rotation invariant method, called radial Tchebichef moments, were proposed by \cite{Mukundan05}.
A comparison between Tchebichef polynomial based moments and Zernike moments was carried out in \cite{Mukundan09A}.
Finally, Mukundan~\cite{Mukundan2014} proposed the Local Tchebichef Moments (LTM), a method specifically developed for texture analysis.
The method is based on the Tchebichef polynomials, and has been modified in order to get local feature representation.
Also, the method uses the relationship between moments to create a histogram in a similar way to what LBP methods do.
%\section{Related Work}
Marcos and Cristobal \cite{MarcosCristobal2013} studied the use of discrete Tchebichef moments for texture classification, comparing it with other methods and using standard datasets (e.g., Outex) to run machine learning experiments.
They concluded that the method was very competitive compared to other feature extraction methods, including the ability to cope with rotation.
However, the discrete Tchebichef moments method is very different than the LTM method proposed in \cite{Mukundan2014}.

In this paper, the objective is to carry out an assessment and to fine tune Mukundan's Local Tchebichef Moments (LTM) method~\cite{Mukundan2014} and also to compare it to various Local Binary Pattern (LBP) based methods.

\section{Local Tchebichef Moments}
For completion, we fully describe the method to compute the LTMs, so results can be easily reproduced.

%The moments can be evaluated using different window sizes (TODO vary window sizes???)
%It is somewhat based on LBP's in the sense that the method produces also an image, and then the histogram of the resulting image is used as features for the analysis or classification.

The method starts by using the orthonormal Tchebichef polynomials of degree $n$ by $t_{n}(x)$.
The following recurrence relation can be used (from \cite{Mukundan04B} and ~\cite{Mukundan2014}):

\begin{equation}
\label{tn}
 t_{n}(x) = \alpha_{1} x t_{n-1}(x) + \alpha_{2} t_{n-1}(x) + \alpha_{3} t_{n-2}(x)   
\end{equation}

where the $\alpha$ can be computed in the following equations, with $x$ from $0$ to $N-1$ and $n$ from $2$ to $N-1$:

\begin{equation}
 \alpha_{1} = \frac{2}{n} \sqrt{\frac{4 n^2 -1}{N^2 - n^2}}
\end{equation}

\begin{equation}
 \alpha_{2} = \frac{(1-N)}{n} \sqrt{\frac{4 n^2 -1}{N^2 - n^2}}
\end{equation}

\begin{equation}
 \alpha_{3} = \frac{(n-1)}{n} \sqrt{\frac{2n+1}{2n-3}} \sqrt{\frac{N^2-(n-1)^{2}}{N^2 - n^2}}
\end{equation}

The initial values for $t$, $t_{0}$ and $t_{1}$, are:

\begin{equation}
 t_{0}(x) = \frac{1}{\sqrt{N}}
\end{equation}

\begin{equation}
 t_{1}(x) = (2x+1-N) \sqrt{\frac{3}{N(N^2-1)}}
\end{equation}

Now the moments can be computed with a convolution mask, with every element $M_{mn}$ calculated as:

\begin{equation}
\label{momentsmn}
 M_{mn}(x,y) = t_{m}(x) t_{n}(y)
\end{equation}

The local Tchebichef moment at a certain location can be computed with the convolution mask created with equation \ref{momentsmn}. 
Mukundan used only convolution masks of size 5x5 in \cite{Mukundan2014}.
That meant computing Tchebichef moments up to order 4, yielding a total of 25 possible moments, computed with each mask from $M_{00}$ to $M_{44}$ (see table \ref{Orders}).

\subsection{Tchebichef Descriptors}
Rather than using the moment values directly, a simple mechanism was devised by Mukundan \cite{Mukundan2014} to mimic the LBP process of creating an image.
Using Lehmer code, Mukundan defined the relative strength of each different order and encoded the order as a certain value.

As he used 5 moments for his experiments, that gave an adequate number of permutations.
The total number of permutations, $5! = 120$ was an adequate compromise ($4!=6$ is too small, and $6!=720$ is too large to fit in the $[0,255]$ interval) in order to create an LTM image with grayscale. %, specially if multiplying the resulting number by 2.
So the LTM image pixels could vary from 0 to 238.
The histogram of the resulting LTM image can be used as the feature set for classification.
However, the number of moments used could be easily extended, with a histogram size beyond the limit of 255 for a grayscale image.

Another interesting decision to be made is to include weights multiplying the local Tchebichef moments, in such a way that it influences the final Lehmer code.
Mukundan showed two possibilities, one with equal weights for every moment, and another with a somewhat arbitrary weight set.
It was not established how the weights influence the classification.

\begin{figure}[!htb]

  \centering
    \includegraphics[width=0.32\linewidth]{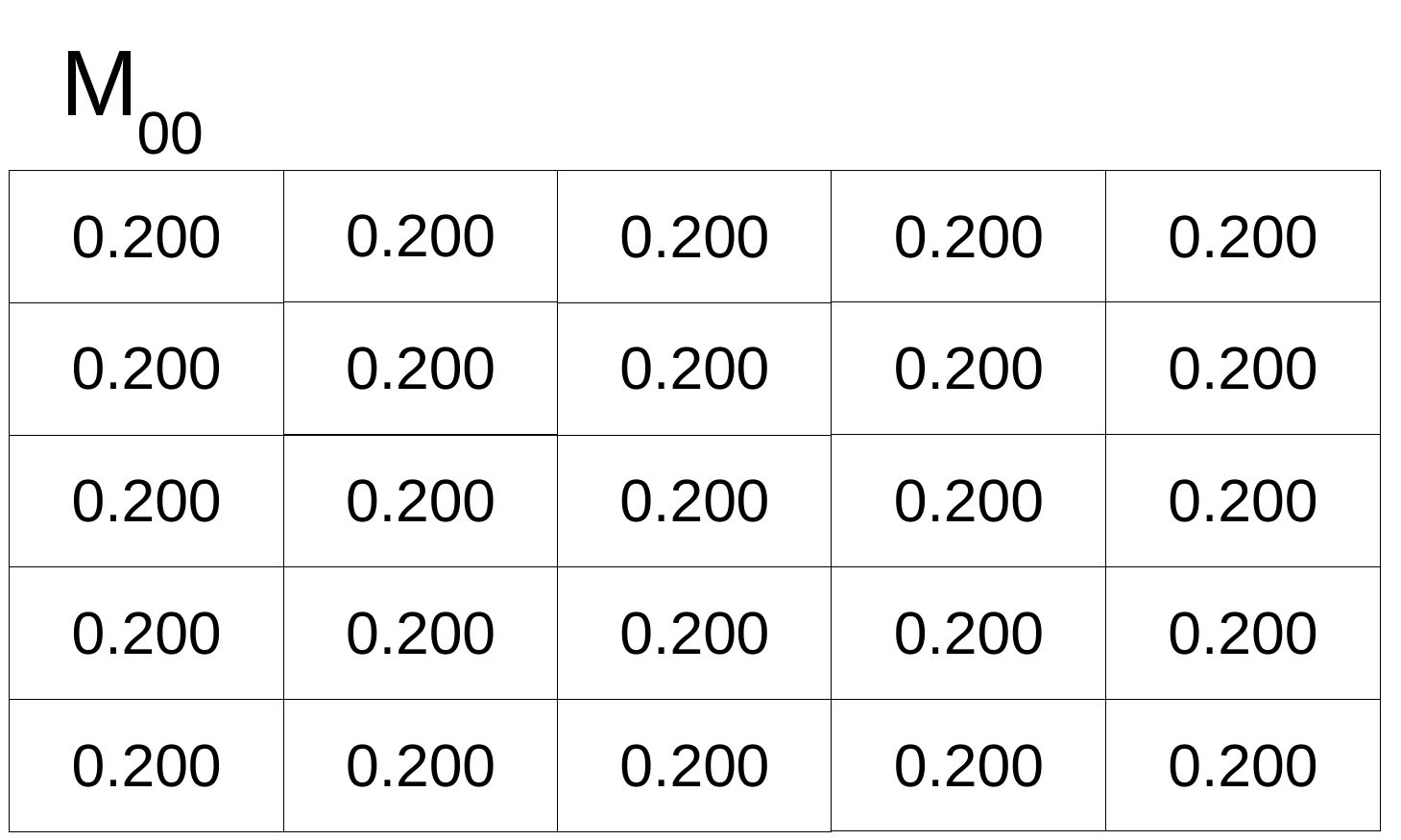}   
  \includegraphics[width=0.32\linewidth]{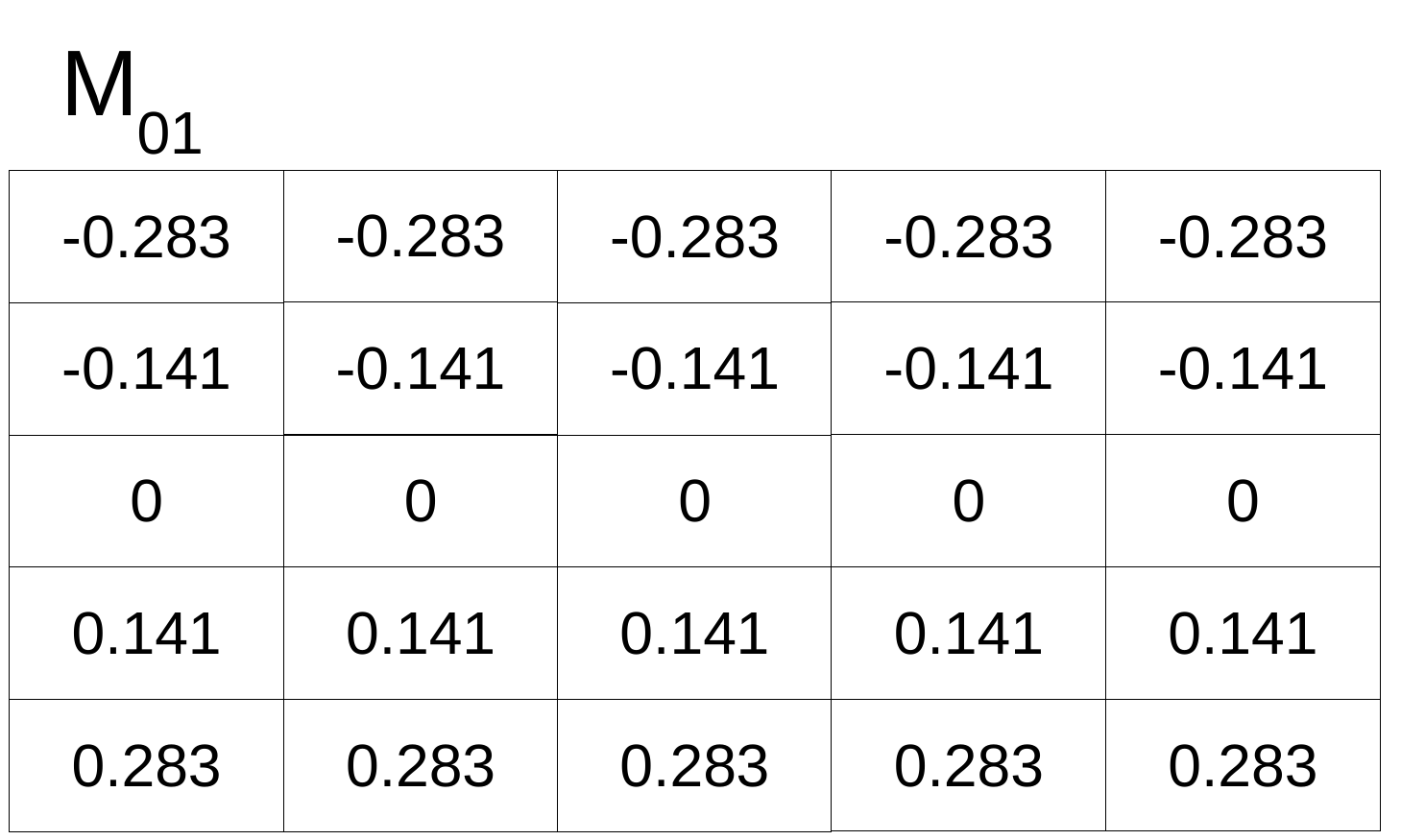}
  \includegraphics[width=0.32\linewidth]{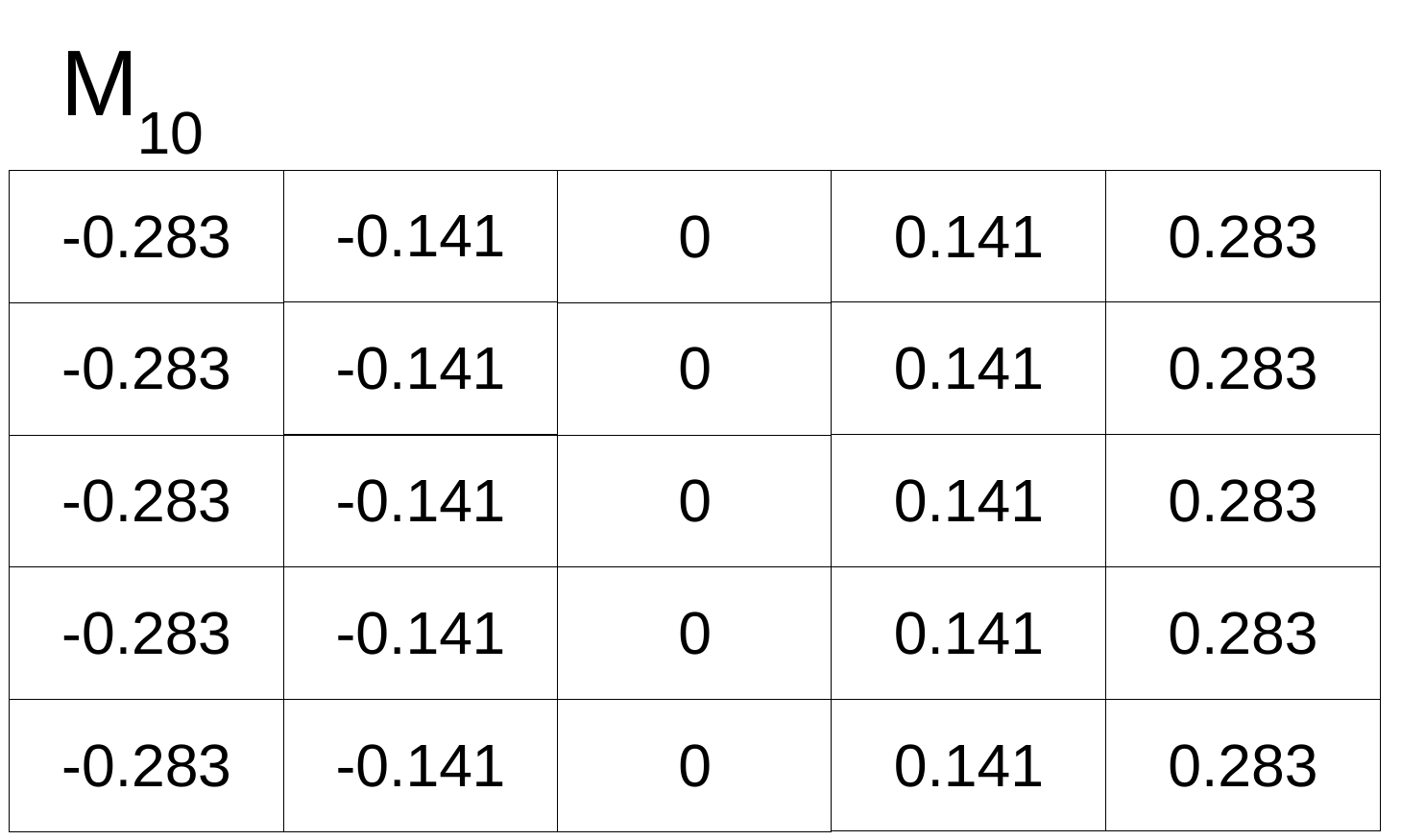}\\
    \includegraphics[width=0.32\linewidth]{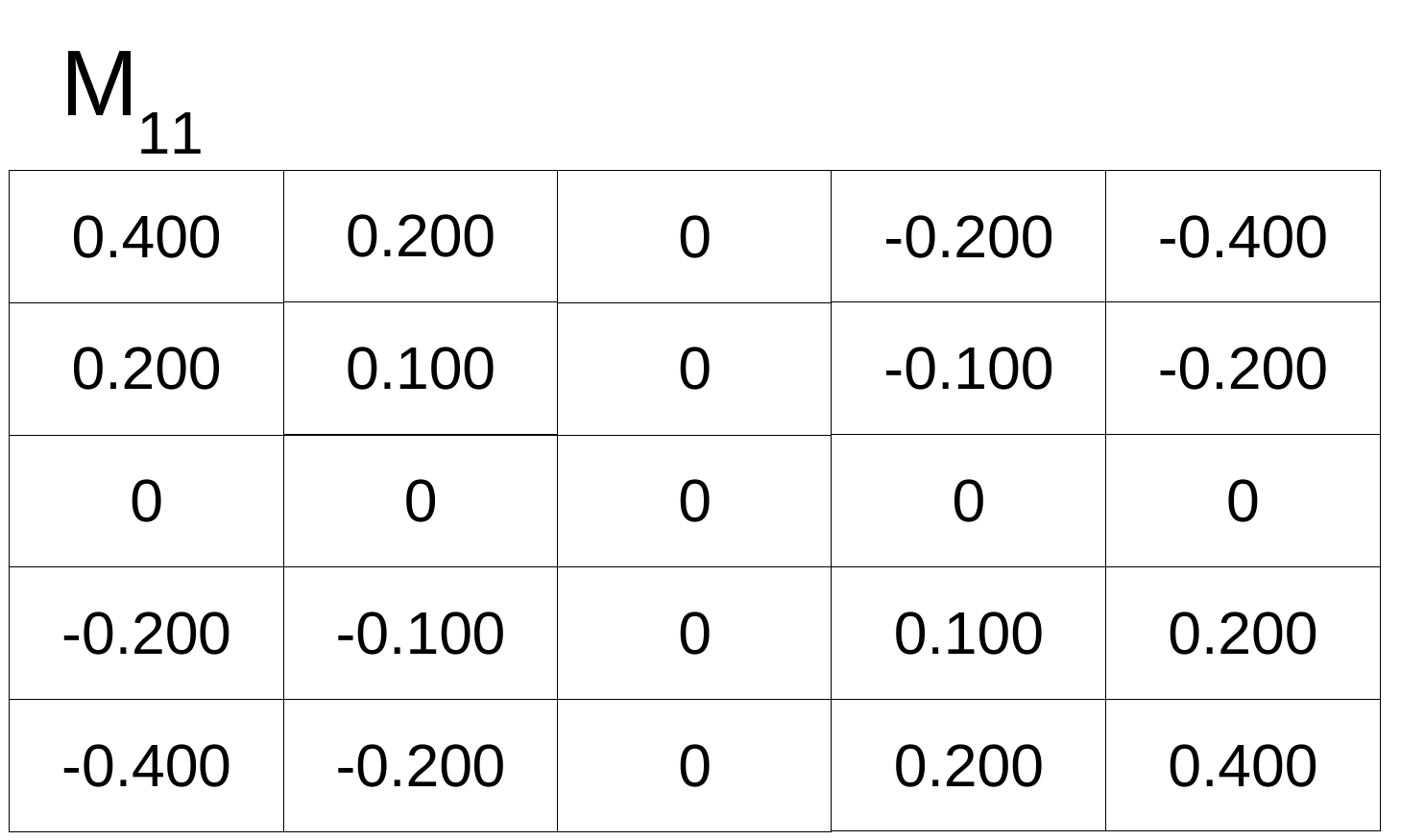}
  \includegraphics[width=0.32\linewidth]{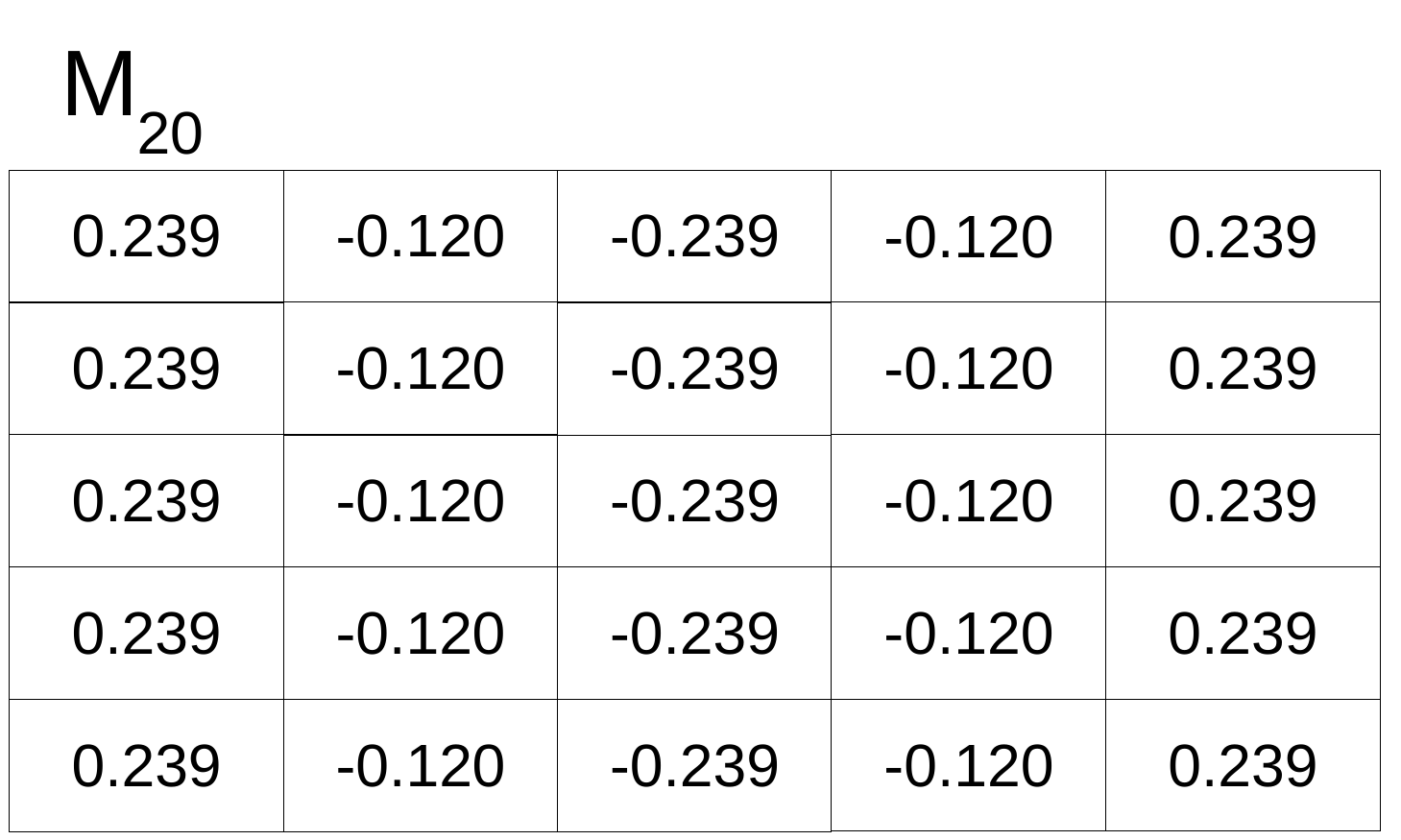}
  \includegraphics[width=0.32\linewidth]{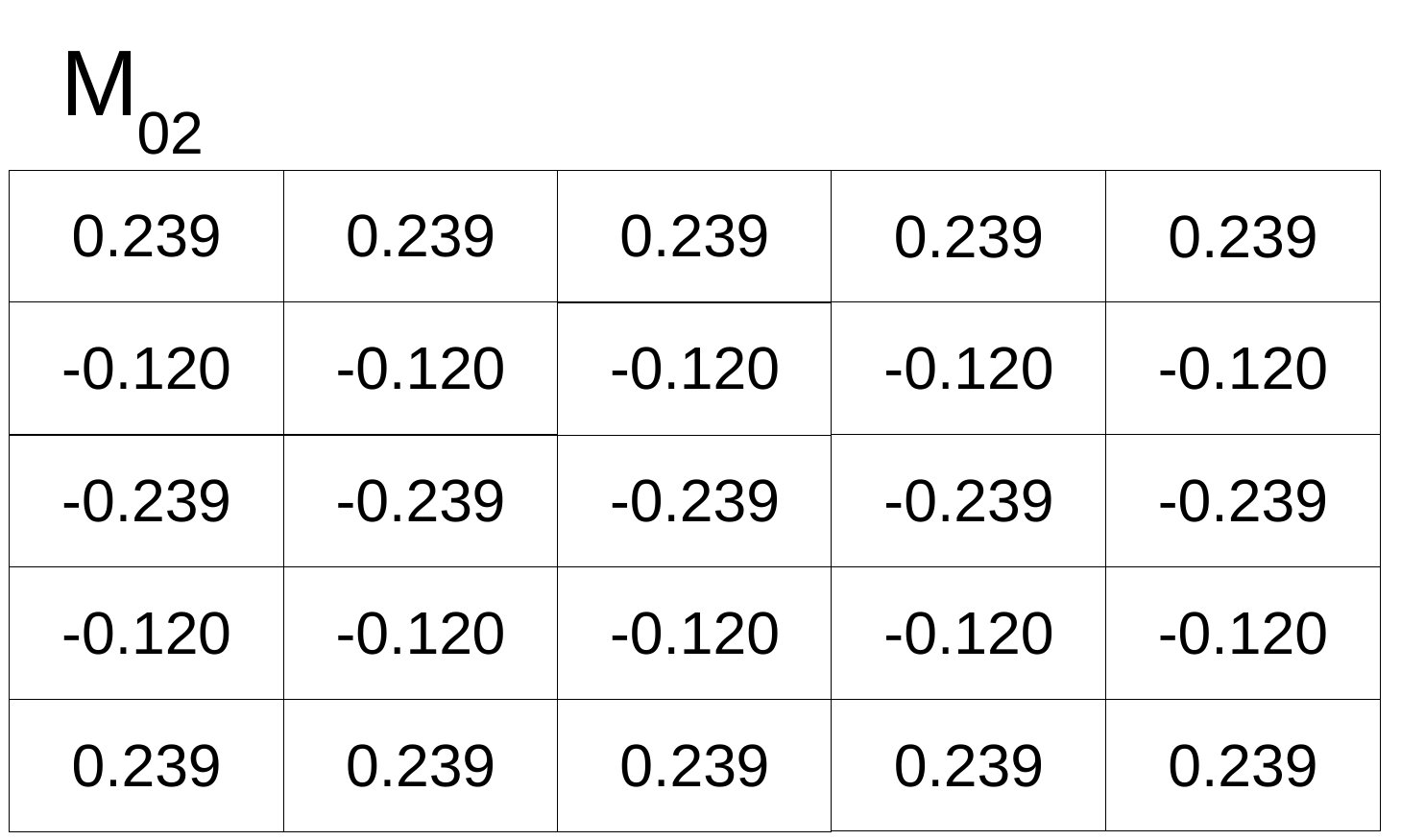}\\
    \includegraphics[width=0.32\linewidth]{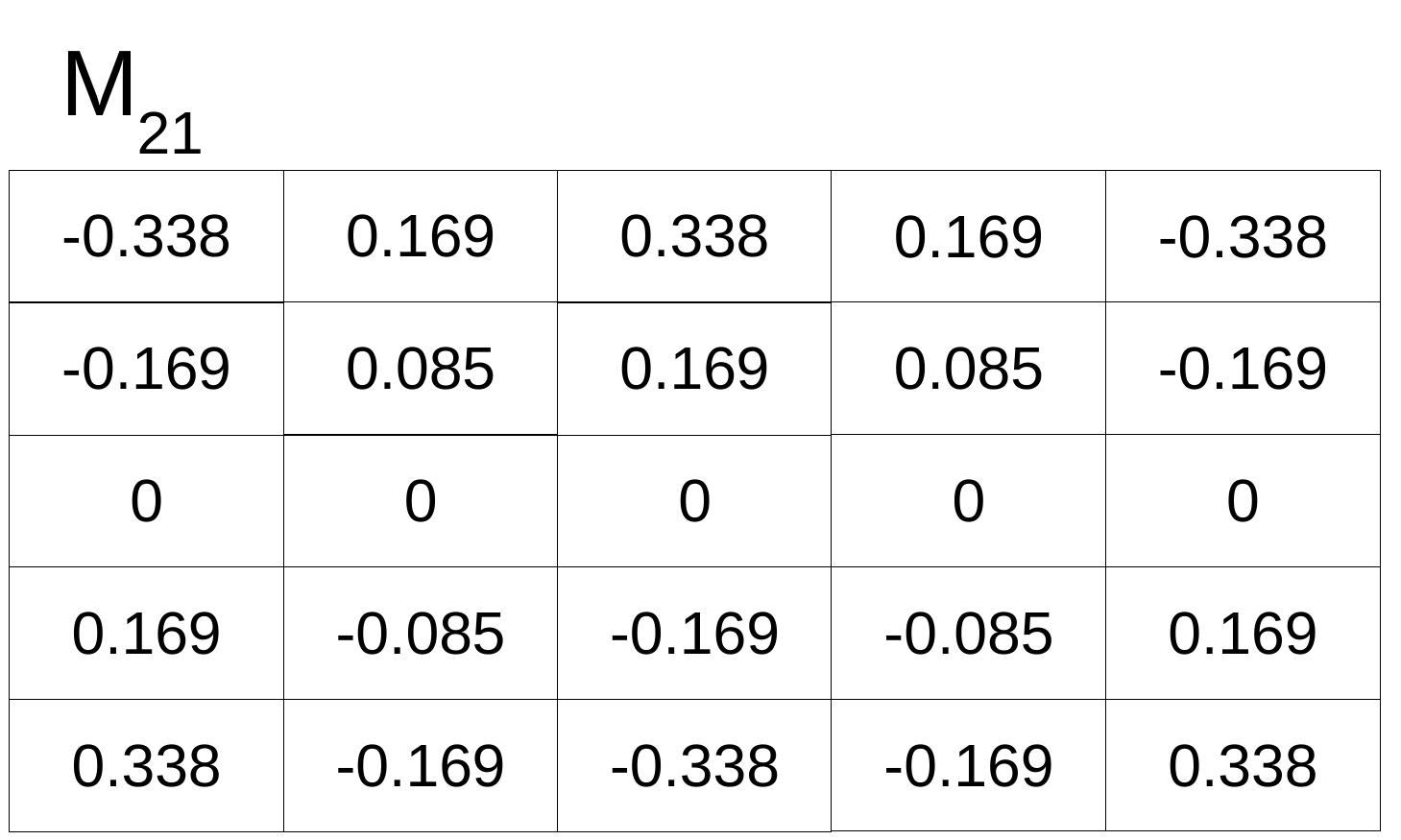}
  \includegraphics[width=0.32\linewidth]{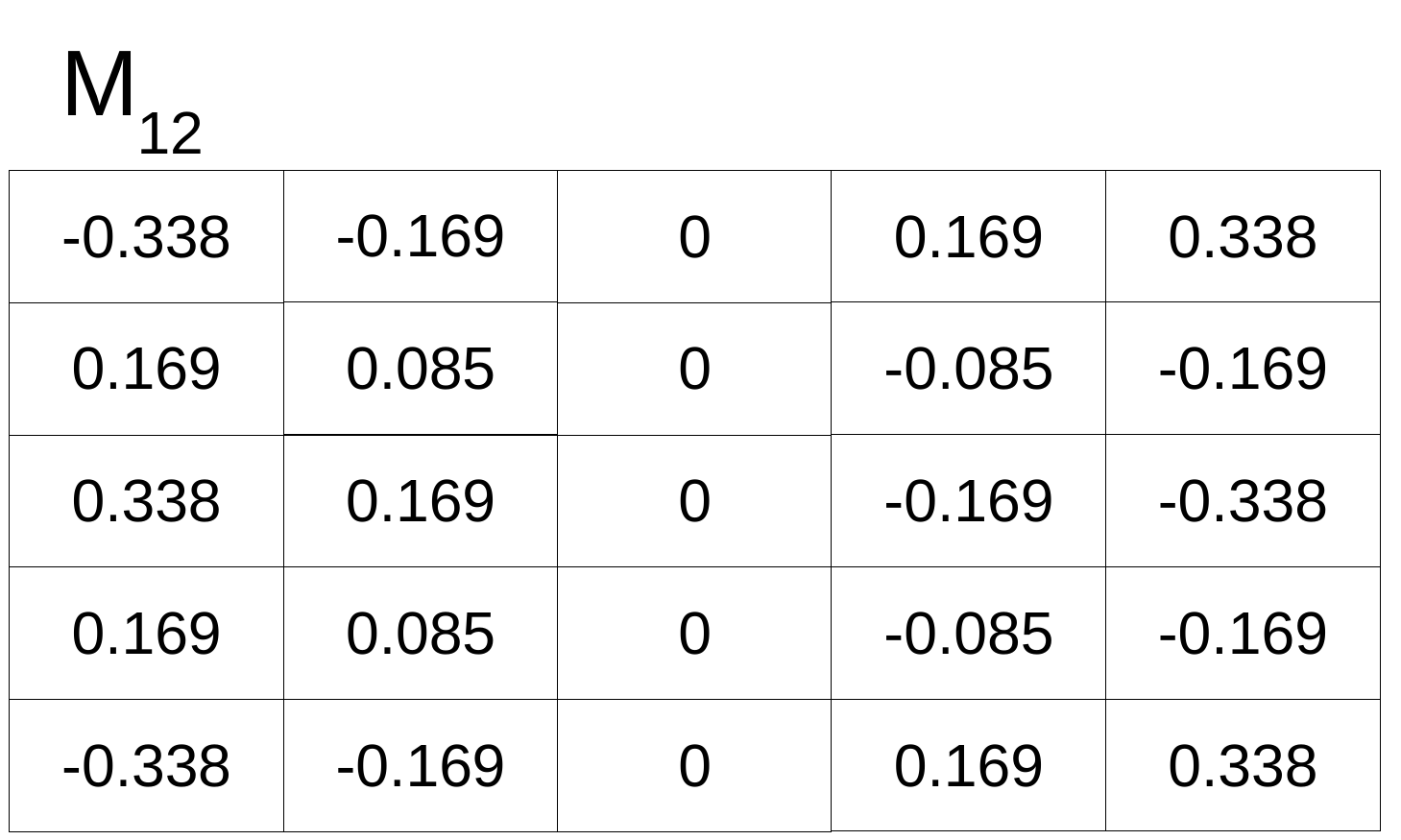}
  \includegraphics[width=0.32\linewidth]{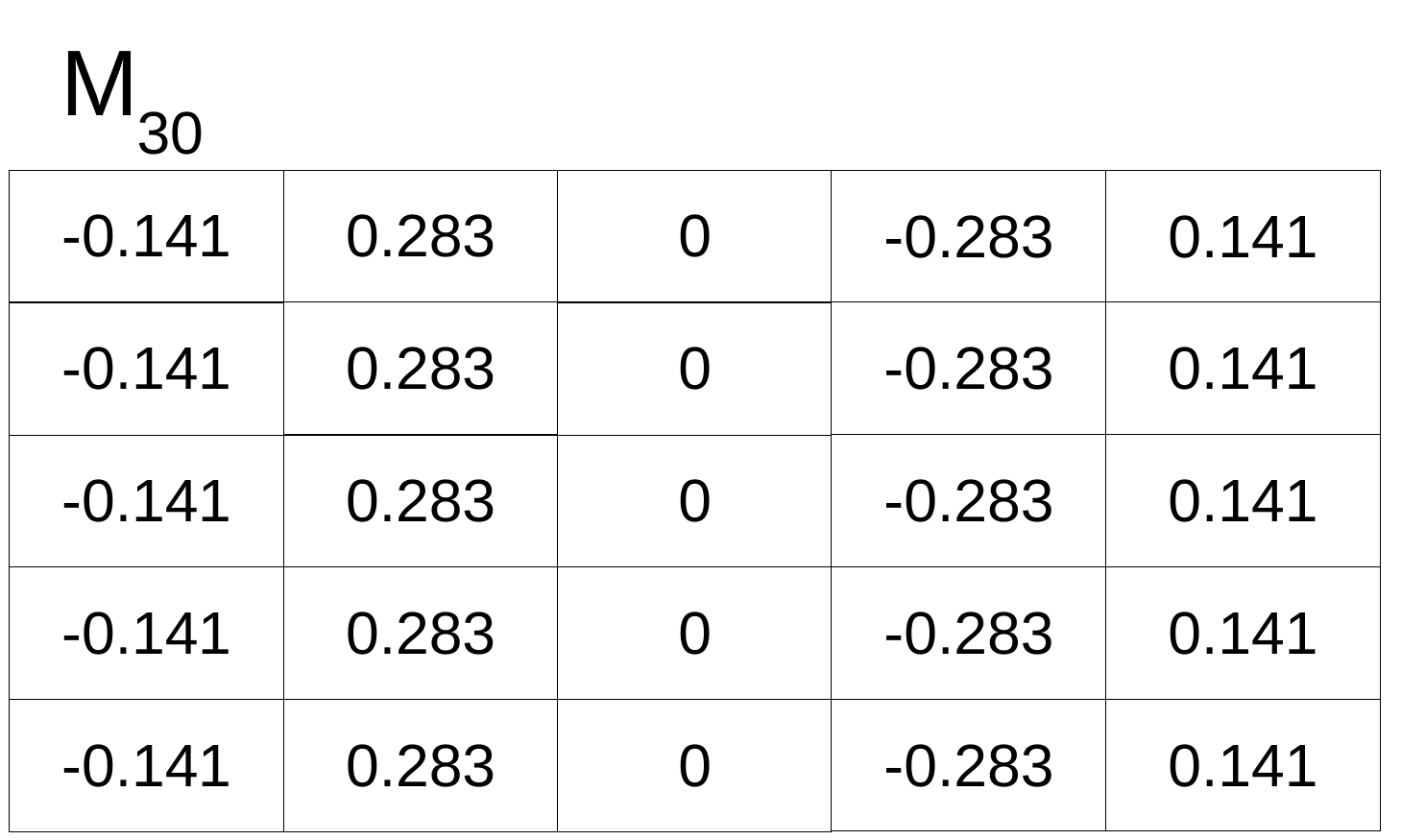}\\
    \includegraphics[width=0.32\linewidth]{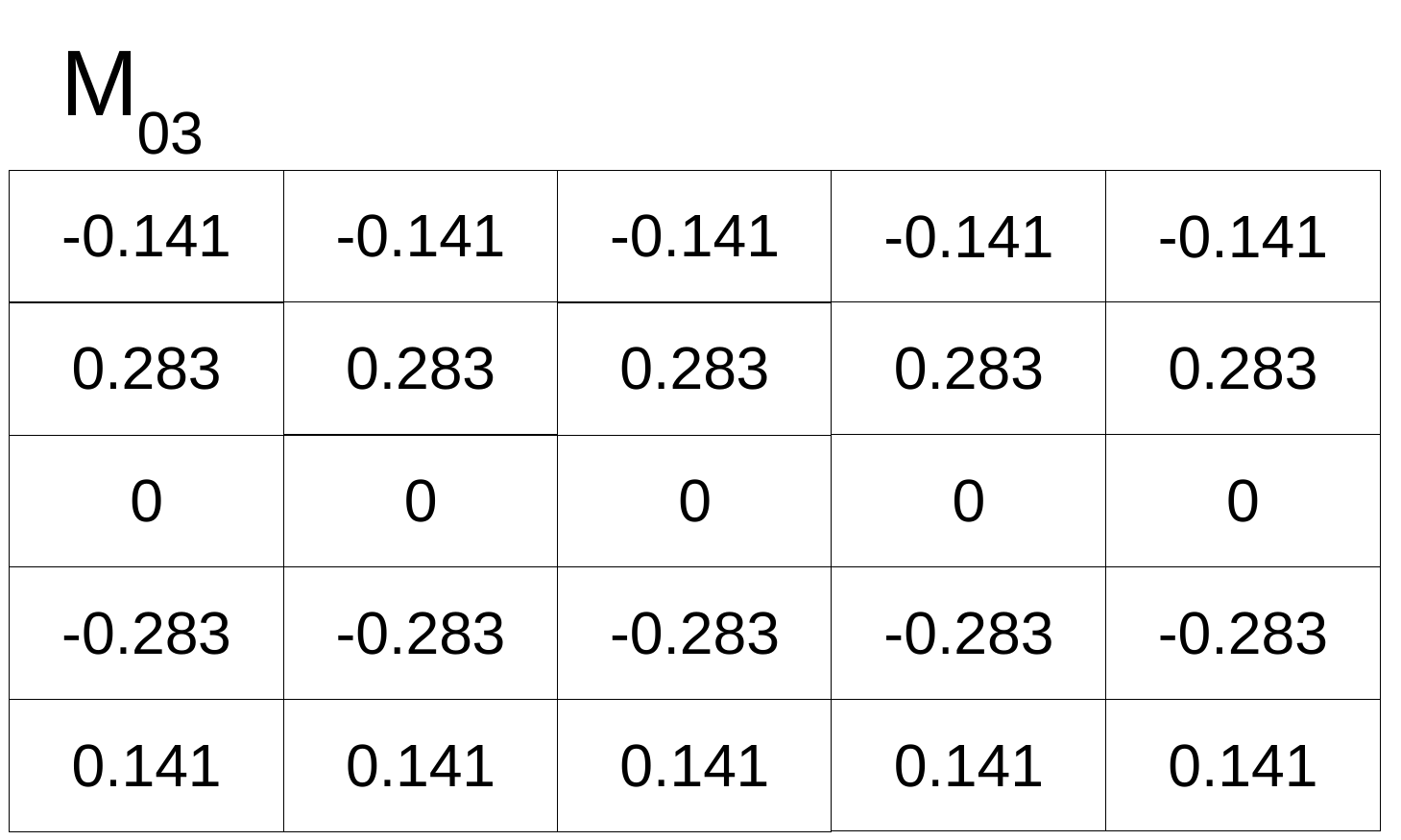}
  \includegraphics[width=0.32\linewidth]{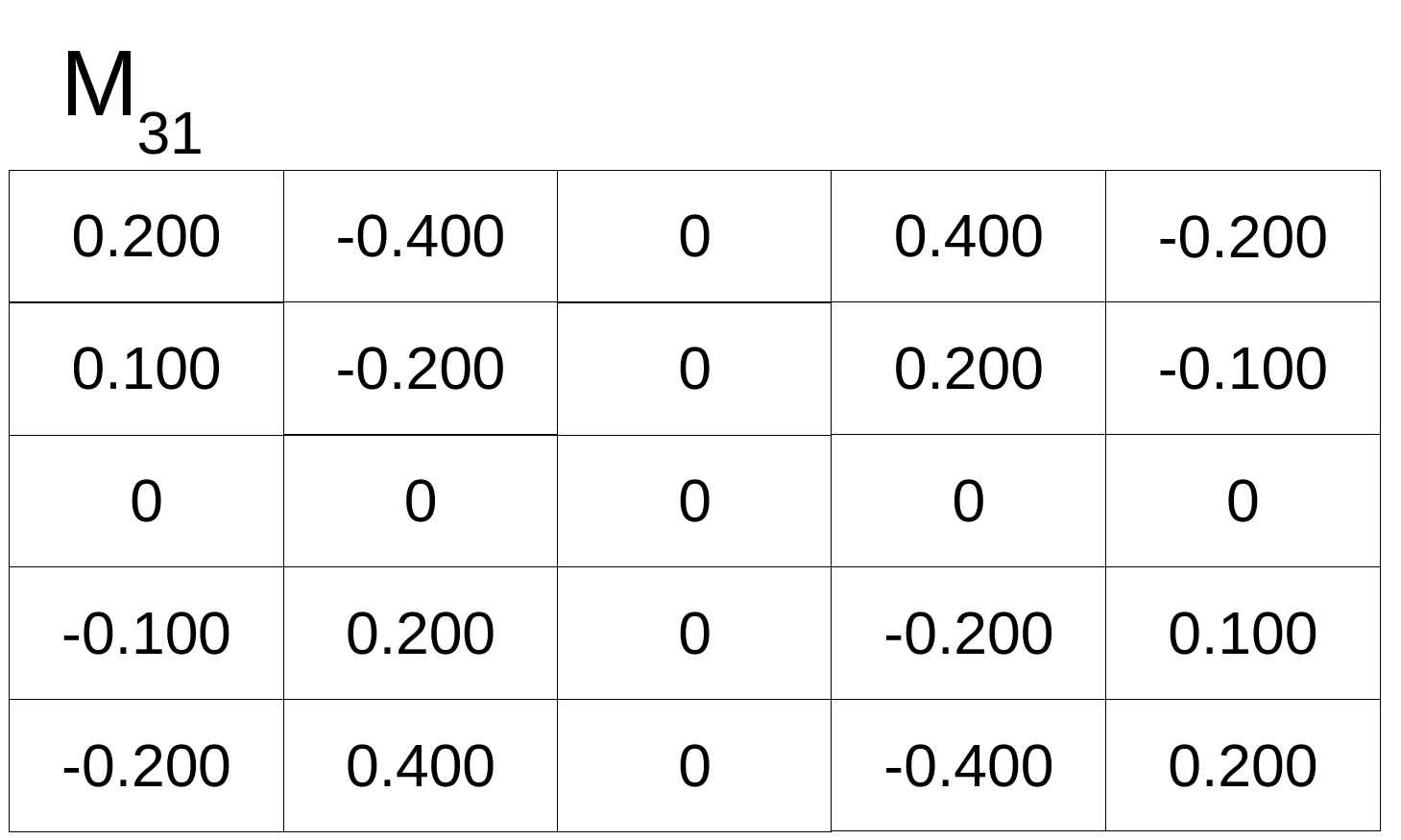}
  \includegraphics[width=0.32\linewidth]{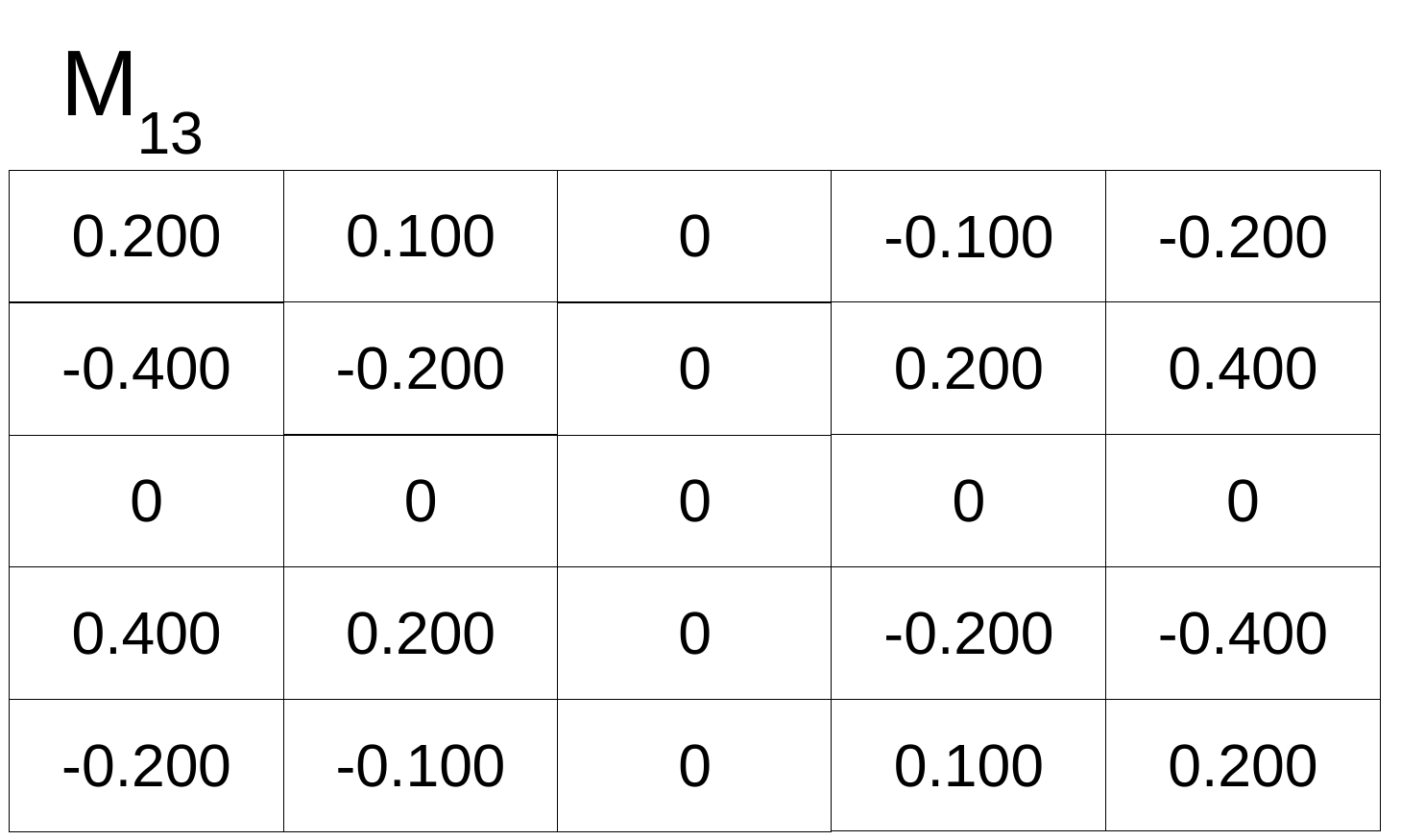}\\
    \includegraphics[width=0.32\linewidth]{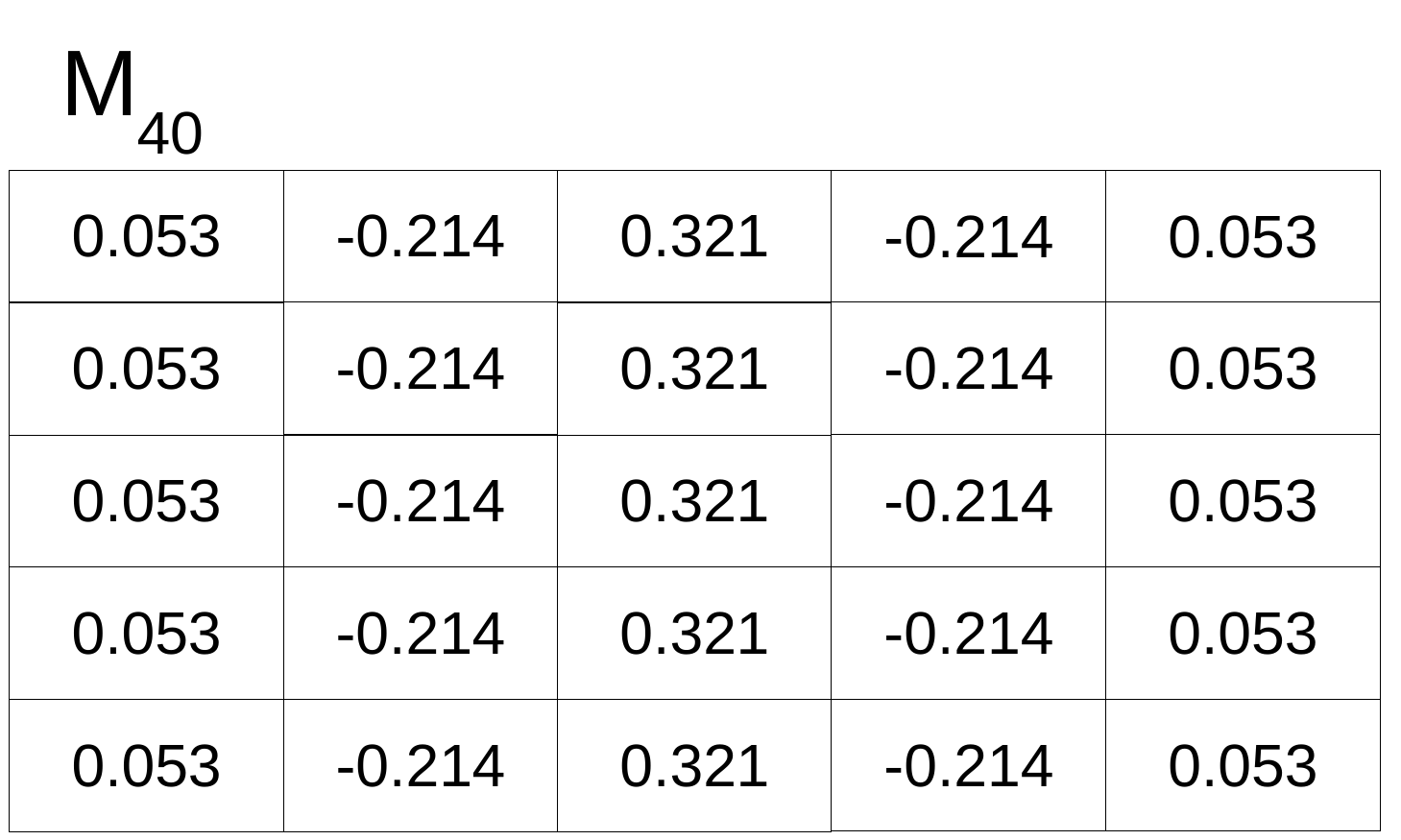}
  \includegraphics[width=0.32\linewidth]{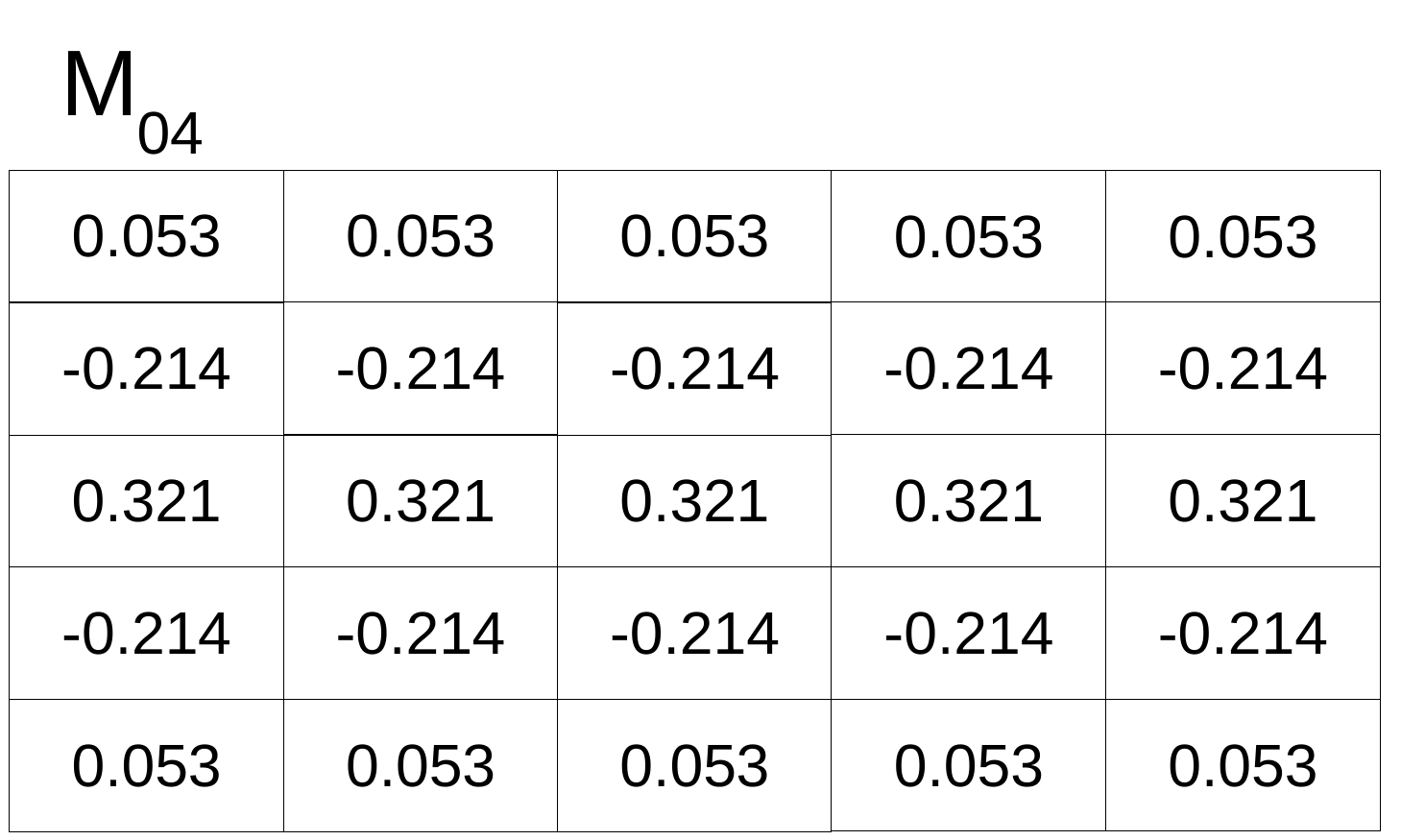}
  \includegraphics[width=0.32\linewidth]{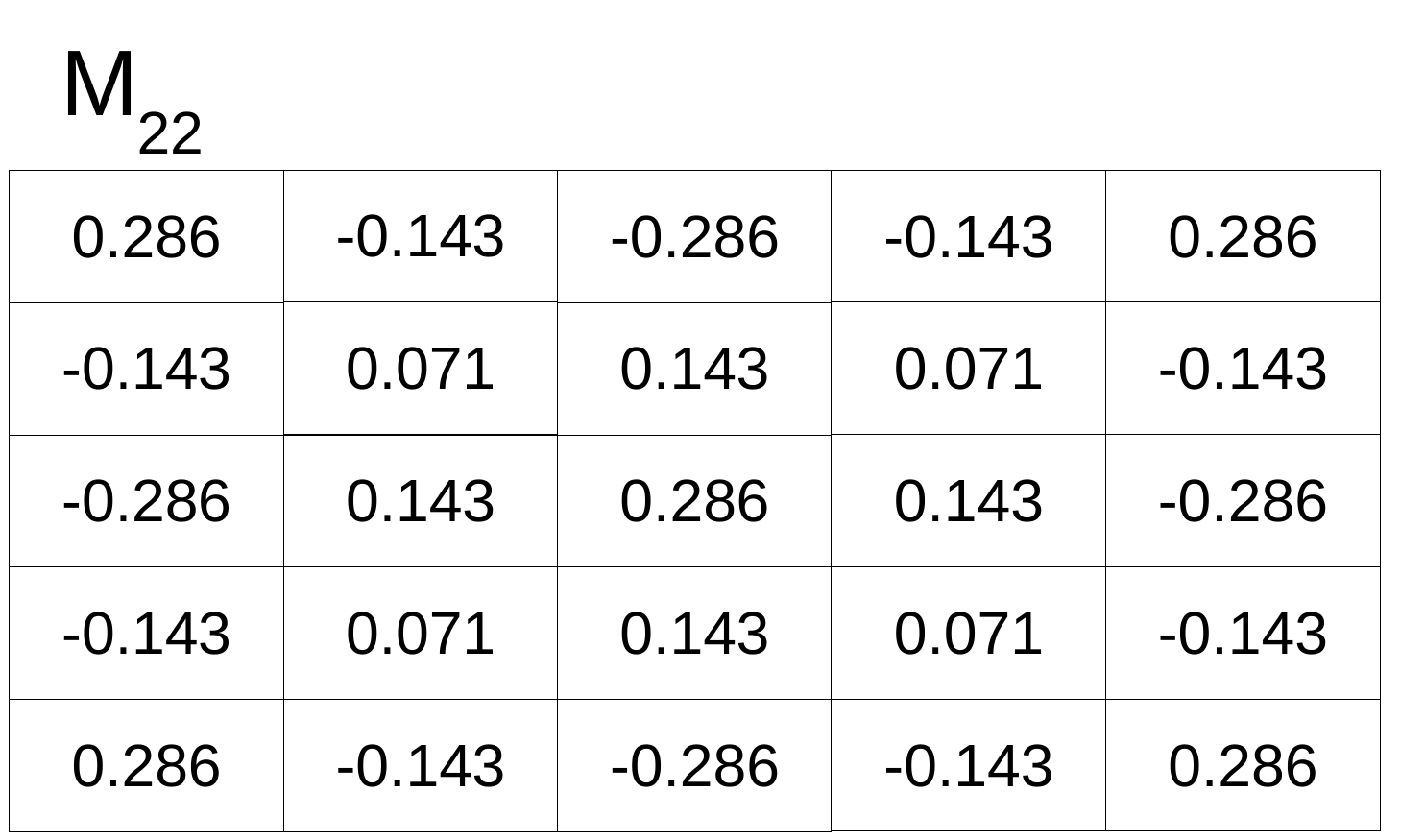}\\
    \includegraphics[width=0.32\linewidth]{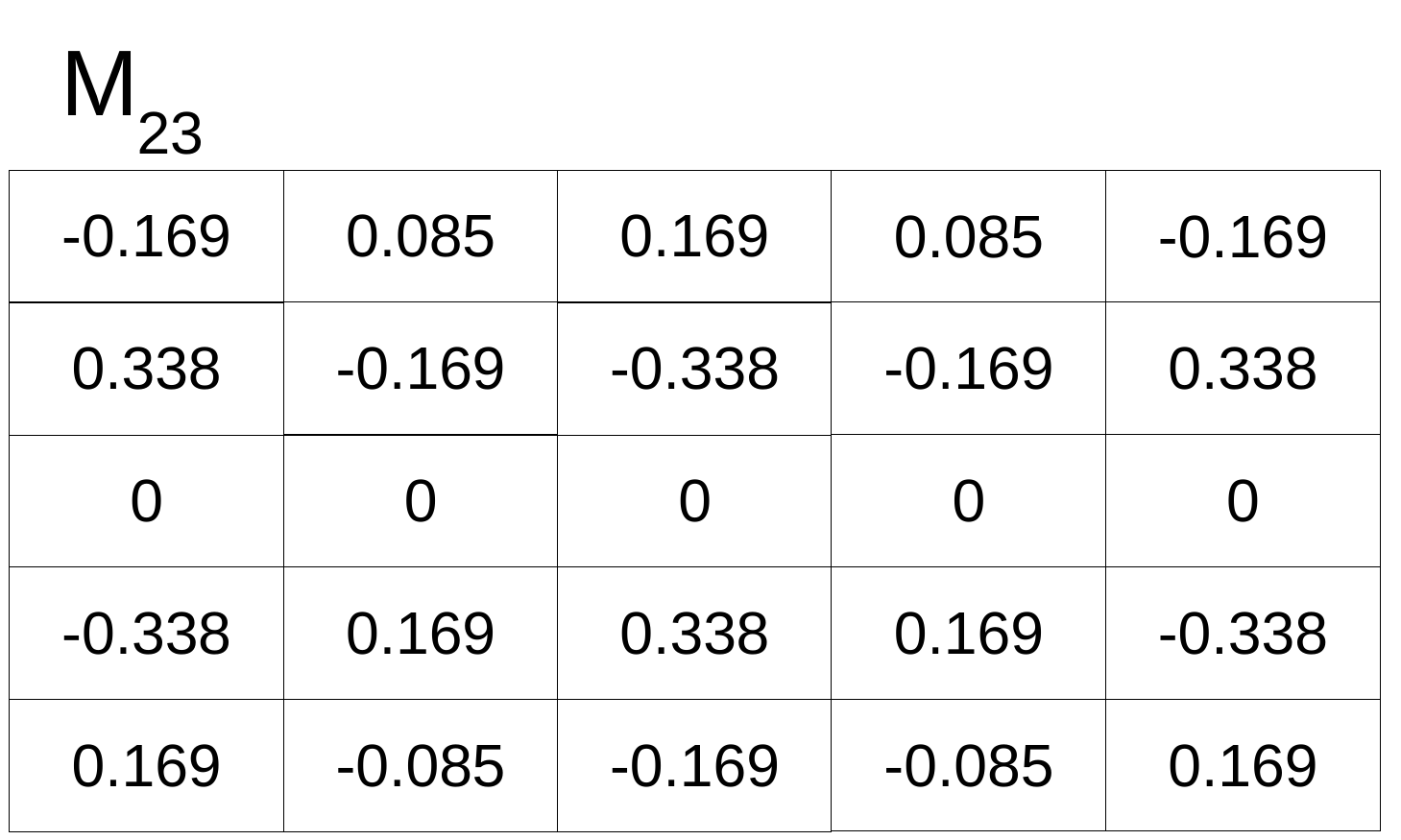}
  \includegraphics[width=0.32\linewidth]{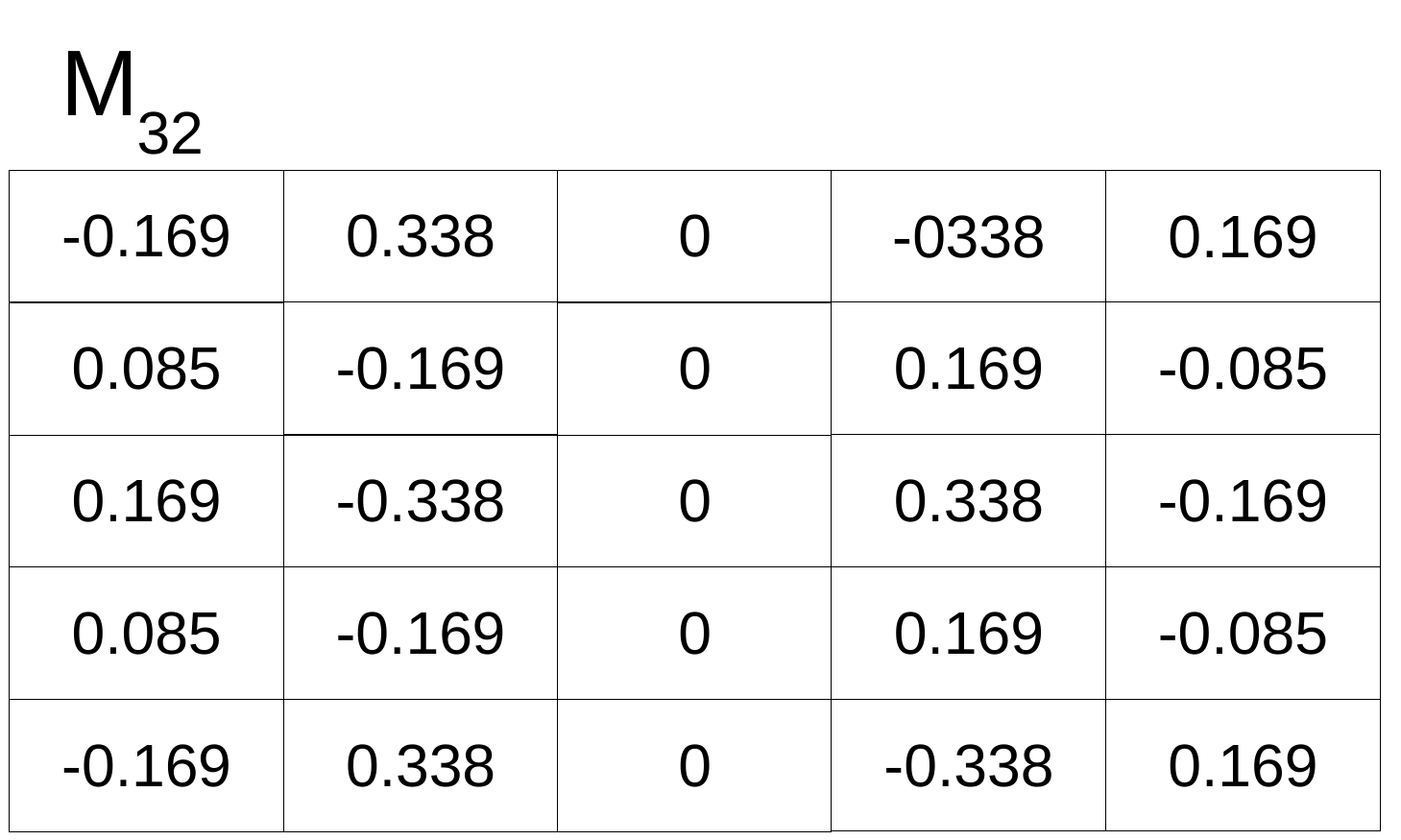}
  \includegraphics[width=0.32\linewidth]{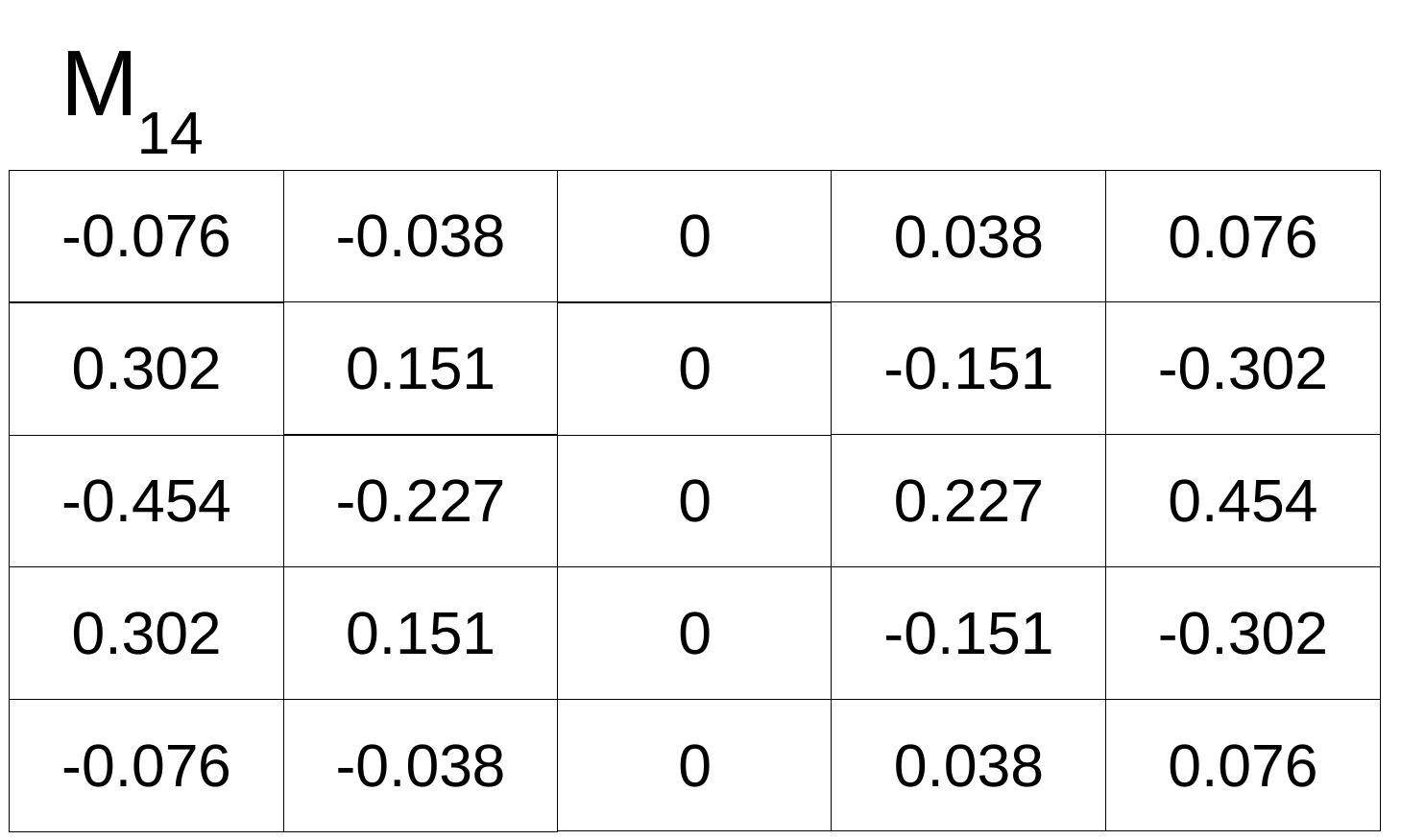}\\
    \includegraphics[width=0.32\linewidth]{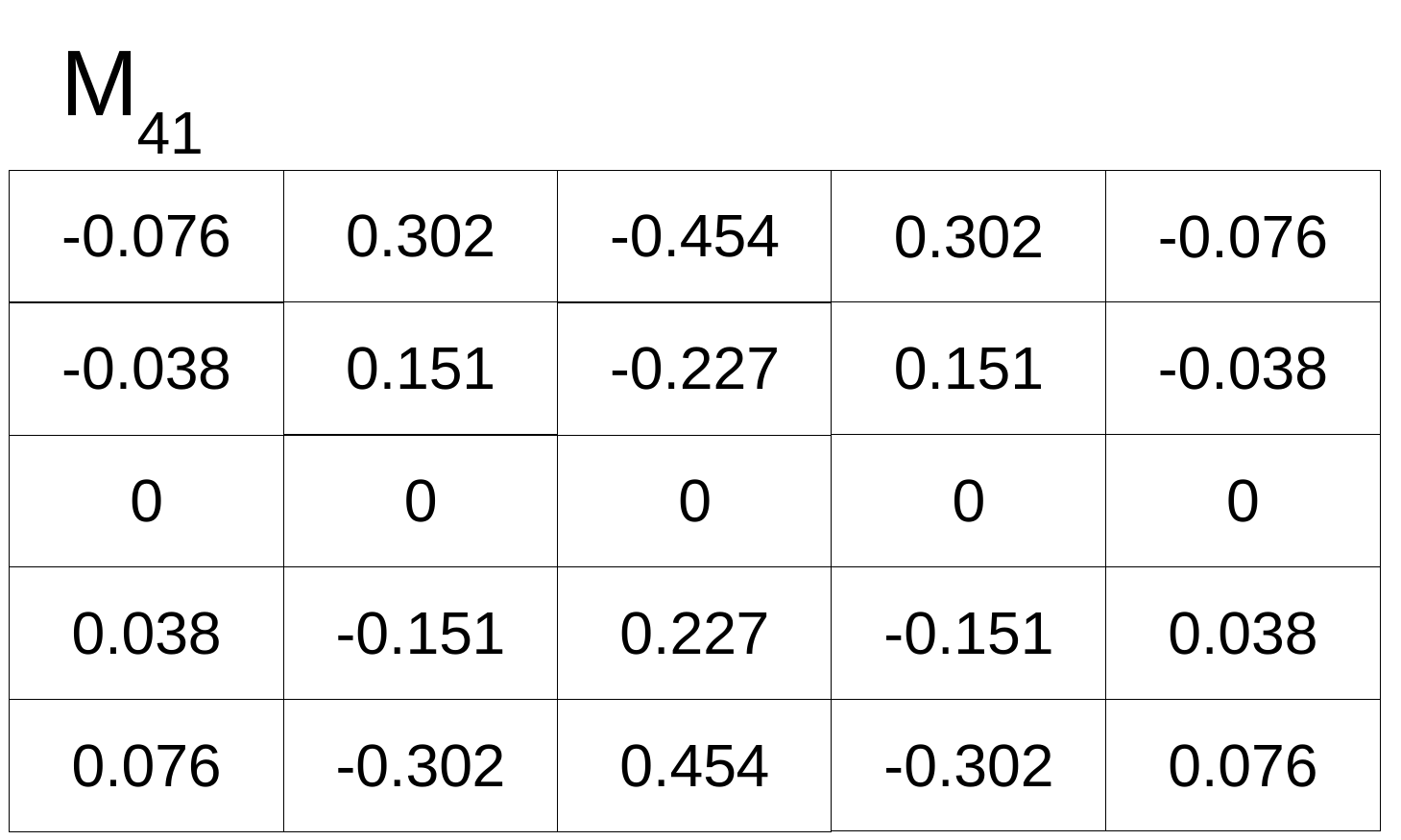}
  \includegraphics[width=0.32\linewidth]{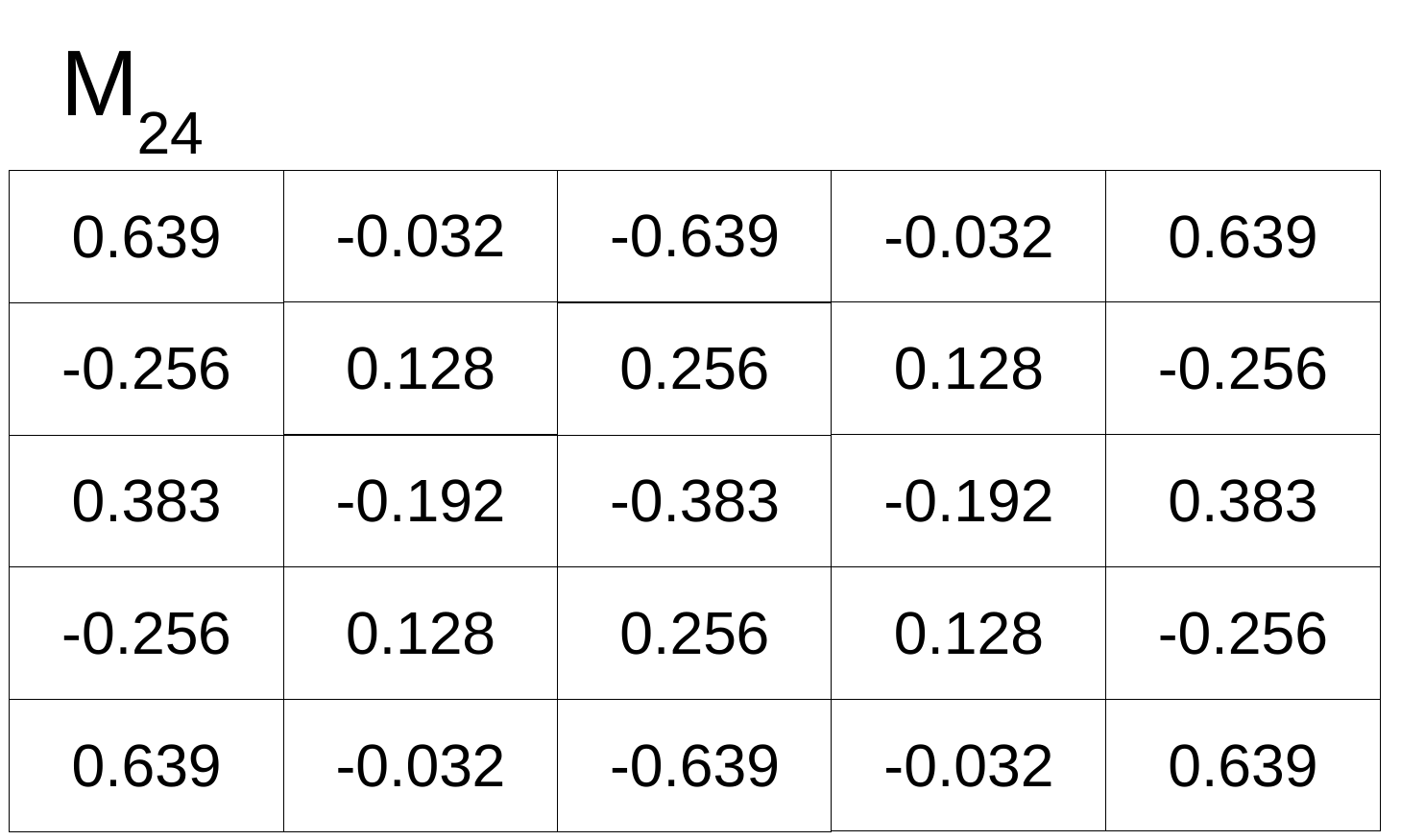}
  \includegraphics[width=0.32\linewidth]{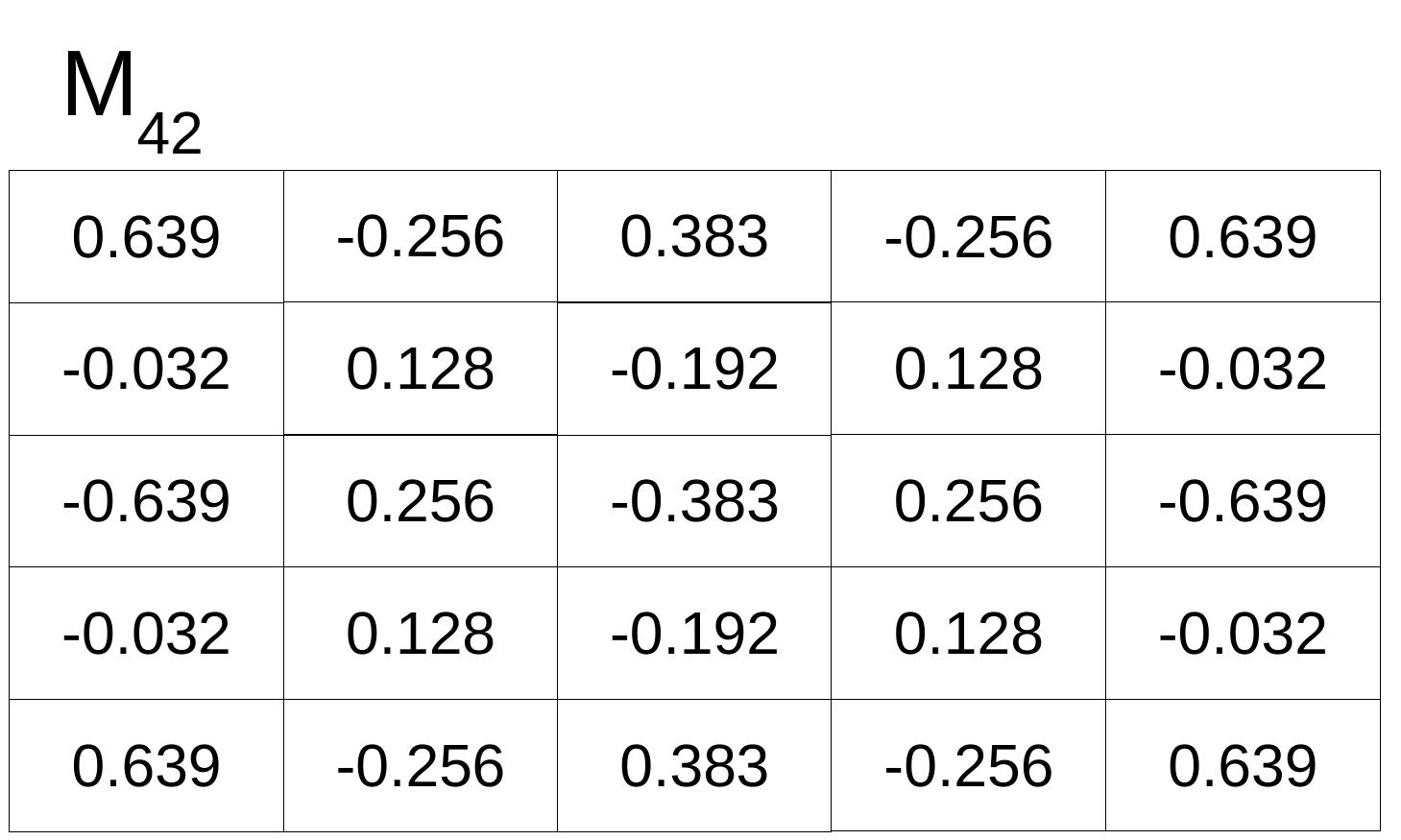}\\
    \includegraphics[width=0.32\linewidth]{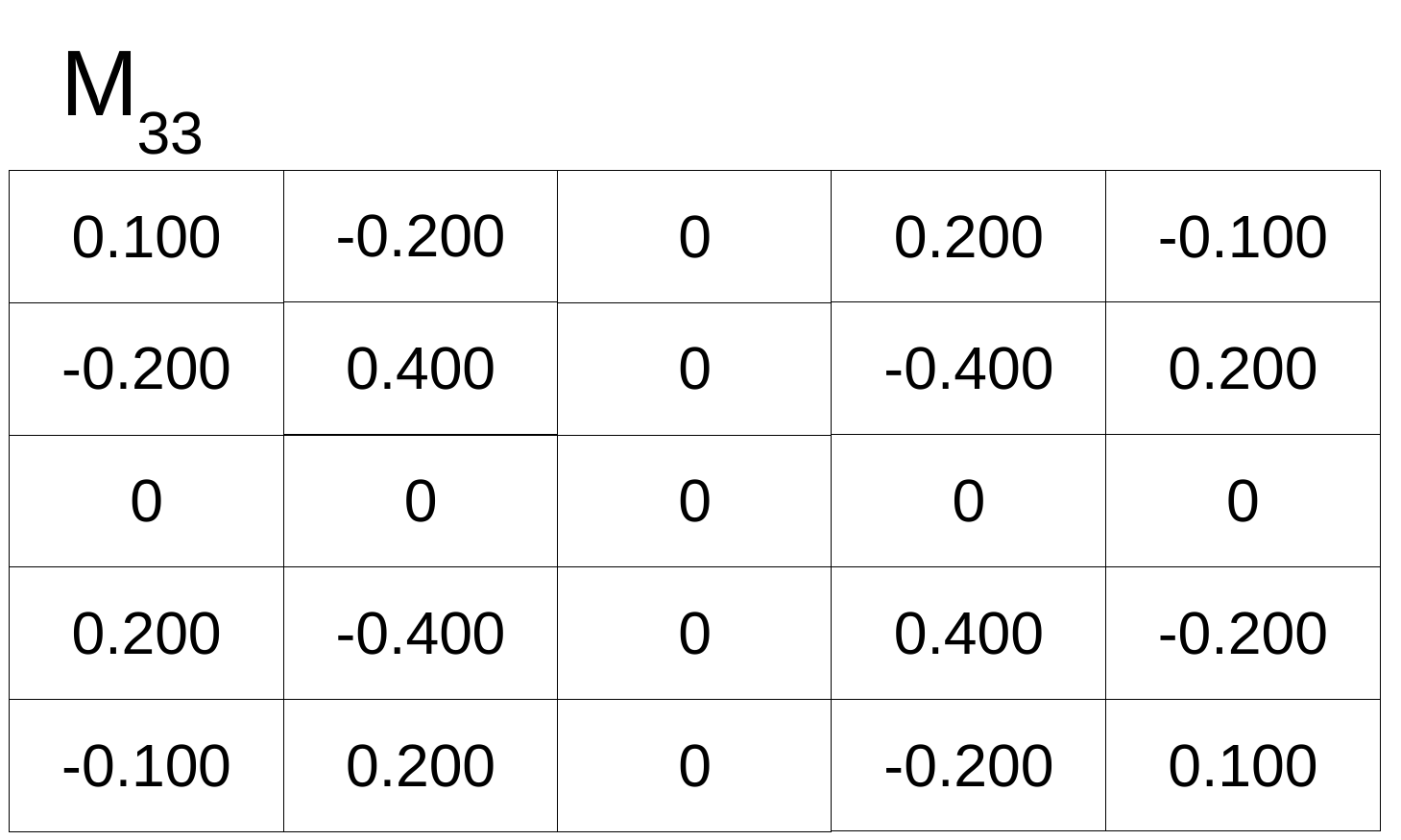}
  \includegraphics[width=0.32\linewidth]{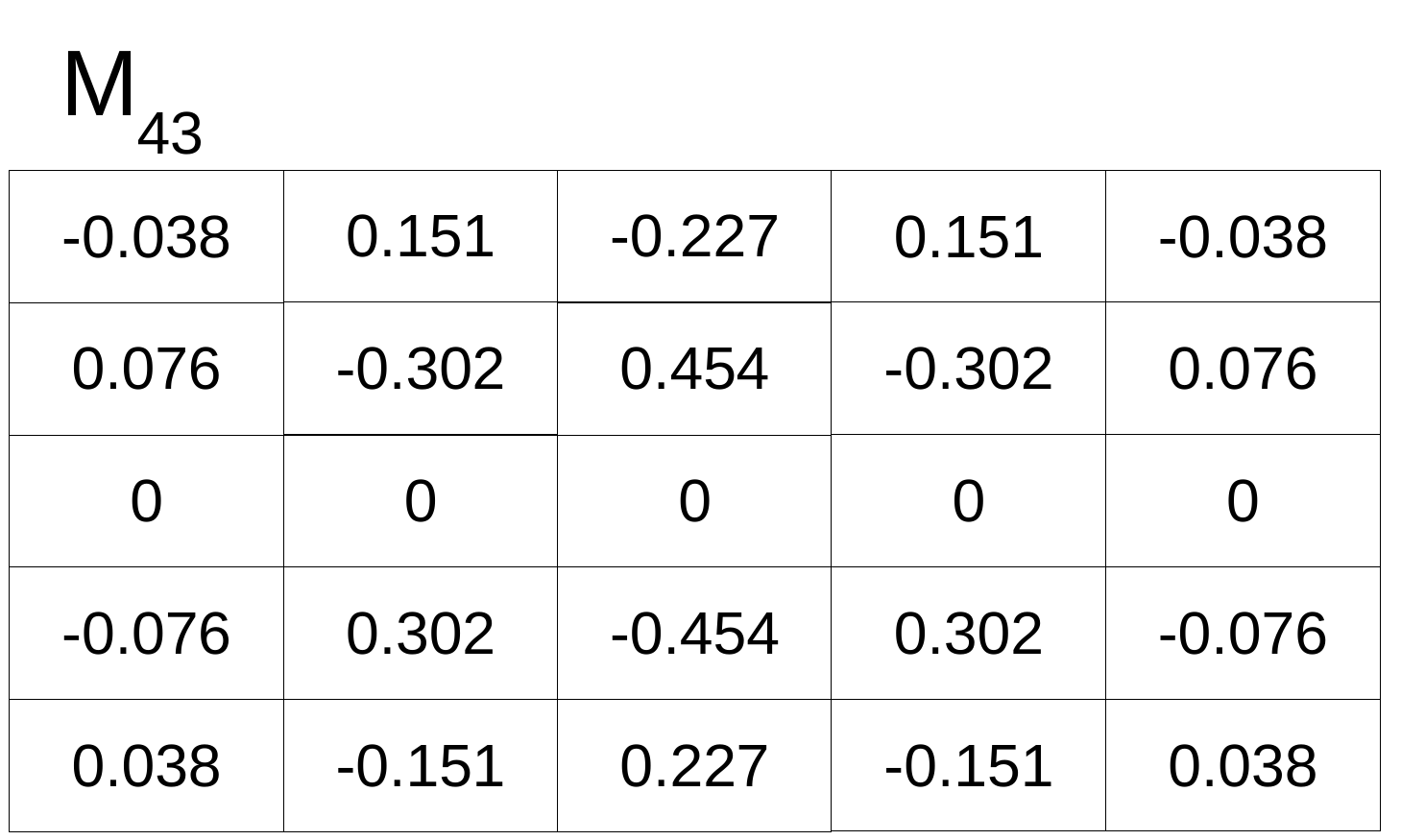}
  \includegraphics[width=0.32\linewidth]{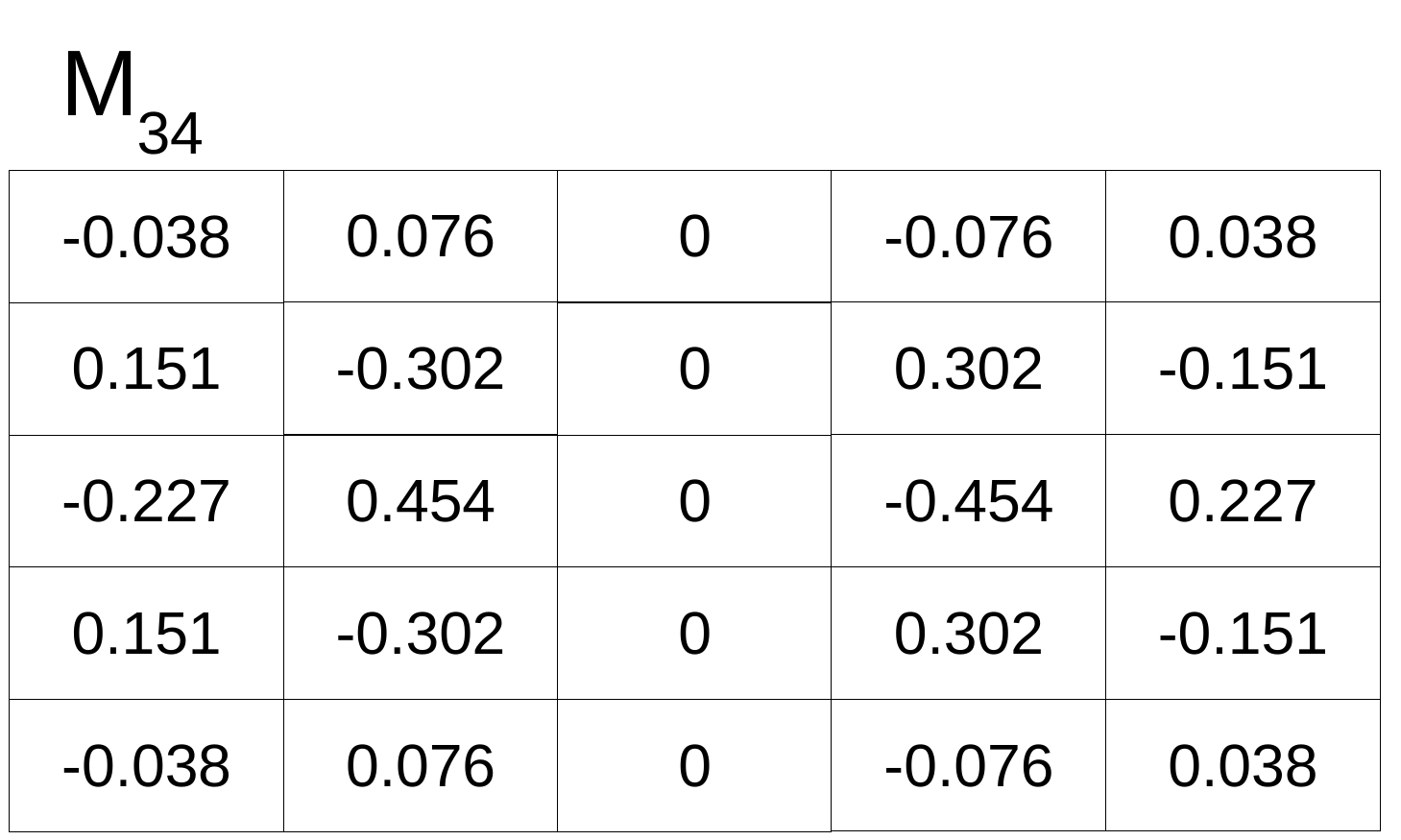}\\
    \includegraphics[width=0.32\linewidth]{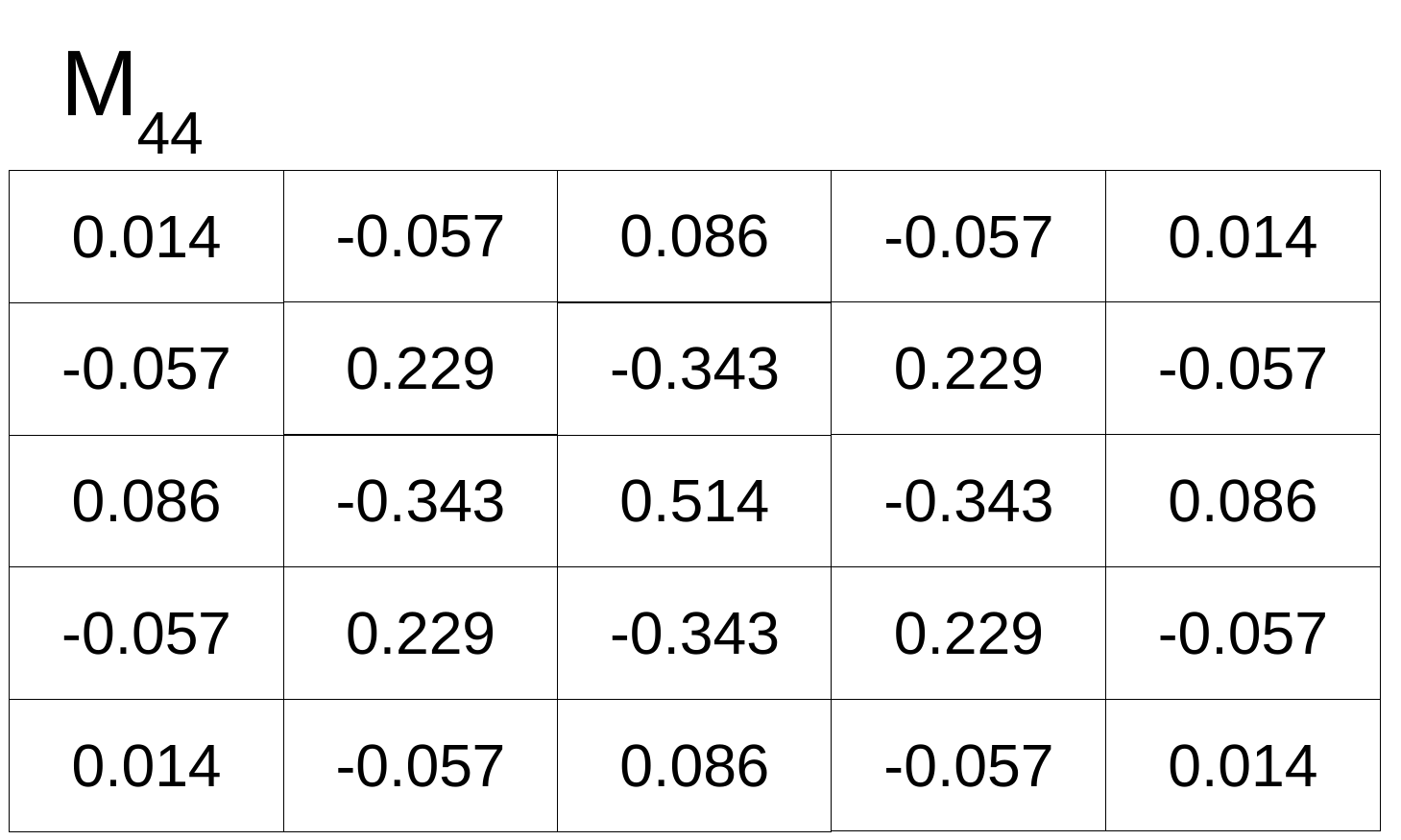}
  \caption{The complete set of 25 convolution masks for the computation of the Local Tchebichef Moments with a kernel size of 5x5. Each value is rounded up to 3 significant digits.}
  \label{25Mpq}
\end{figure}

To complete the description of this section, figure \ref{25Mpq} shows the 5x5 convolution masks for the 25 possible order combinations.
The figures can be used to implement the LTM computations directly, without using equations \ref{tn} to \ref{momentsmn}.

%TODO: Figure 5x5 convolution masks

\clearpage

\begin{figure}[!htb]

  \centering
    \includegraphics[width=\linewidth]{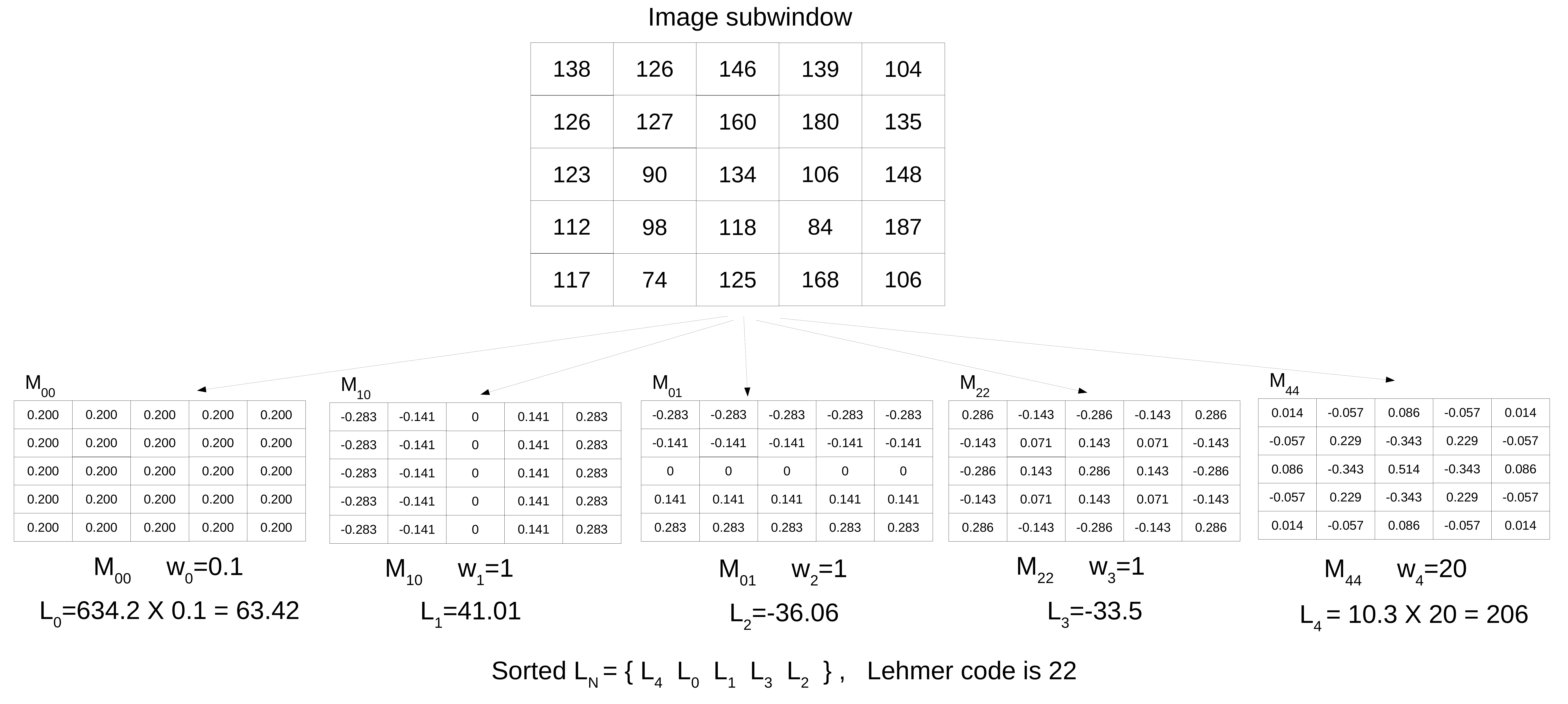}   
    \caption{An example of the computation of a pixel in the LTM image. The centre pixel is computed using weights, similar to the example in \cite{Mukundan2014}.
The $L_{N}$ values are computed using the convolution matrices shown in the figure, and multiplied by the weights. The final value for the Lehmer code is shown, this is the pixel value of the resulting LTM image at the position x=2, y=2.}
\label{LTMcomputation}
\end{figure}

As an example, figure \ref{LTMcomputation} shows the computation of the LTM and its corresponding Lehmer code (to account for the specific permutation of the factors $L_{N}$).
In this example we used $M_{00}$, $M_{11}$, $M_{22}$, $M_{33}$ and $M_{44}$, with weights 0.1,1,1,1,20 respectively.
The final answer for the central pixel in the 5x5 image is the Lehmer code of the permutation (order) of the $L_{N}$ factors.
Along the image, many permutations will happen, so an image can be created with the Lehmer code values and a correspondent histogram can be computed as well.

%TODO Figure similar to fig 6.7 Mukundan2014

\begin{figure}[!htb]
   \centering
     \includegraphics[width=0.32\linewidth]{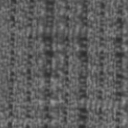}   
     \includegraphics[width=0.32\linewidth]{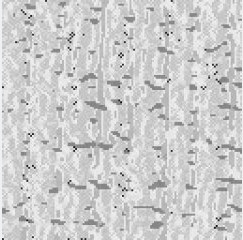}
     \includegraphics[width=0.32\linewidth]{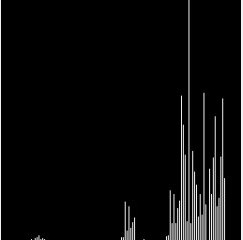}
     \caption{An image created with the LTM method using weights $.1,1,1,1,20$ for one of Outex21 images. The first image on the left is the original texture, the middle image is the LTM results and the last one is the histogram (stretched to up to the bin 239).}
 \label{LTMfigure1}
\end{figure}

\begin{figure}[!htb]
  \centering
    \includegraphics[width=0.32\linewidth]{000002.jpg}   
    \includegraphics[width=0.32\linewidth]{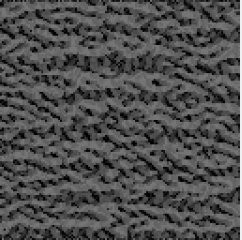}
    \includegraphics[width=0.32\linewidth]{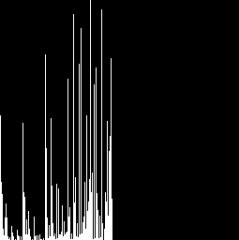}
    \caption{An image created with the LTM method using equal weights of 1 for same Outex21 image as in figure \ref{LTMfigure1}. The first image on the left is the original texture, the middle image is the LTM results and the last one is the histogram, with only 120 bins.}
\label{LTMfigure2}
\end{figure}

In figures \ref{LTMfigure1} and \ref{LTMfigure2} two examples of texture images getting the feature extraction are shown.
In the first example, the histogram was stretched by multiplying the final Lehmer code value by 2, using bins from 0 up to 238.
In the second example, a different combination of moments and weights were used, and the histogram was left with the original number of bins (120, or from 0 to 119).
These two figures demonstrated that there are many different ways of extracting the information from an image using the LTM method.
Of course that for the purpose of machine learning, there is no need to stretch the histograms in such a way that was used in figure \ref{LTMfigure1}. 
This is done purely for visualisation purposes.
One can just use each bin of the histogram as one features, so the feature space has as many dimensions as the histogram's number of bins.


%\clearpage

\section{Comparing LTM with LBP: Preliminary Results}
%\section{Preliminary Results}
Initially we wanted to experiment with the weights and with the usage of different combinations of orders (some arbitrary combination of $M_{pq}$).
The experiments used 4 subsets of the Outex \cite{OjalaEtAl2002} dataset seen in table \ref{OutexTable}.
Two of the datasets have test images that were rotated.
%TODO mention that the symmetric kernels may be rotation invariant too

\subsection{LBPs}
Local Binary Patterns (LBP) was proposed by Ojala et al.~\cite{OjalaEtAl2002}, and it has been modified to cater for special applications over the years.
Silva et al.~\cite{SilvaEtAl2015} did an extensive work in comparing these different LBPs.
A library was made available by Silva et al.~\cite{SilvaEtAl2015} for about 11 different variations of LBP based feature extraction. %(TODO does not matter, only use a few anyway)
In this paper we used 4 of them, namely:
\begin{enumerate}
\item the original LBP by Ojala et al. ~\cite{OjalaEtAl2002}
\item CS-LBP by Heikkila et al. ~\cite{HeikkilaEtAl2006}
\item CS-LDP by Xue et al. ~\cite{XueEtAl2011}
\item XCS-LBP by Silva et al.~\cite{SilvaEtAl2015}
\end{enumerate}

\subsection{Datasets and the Set Up of the Tchebichef Feature Extractor}

\begin{table}[]
\begin{center}
\caption{Outex datasets}
\label{OutexTable}
\begin{tabular}{|l|c|c|c|c|}
\hline
         & \begin{tabular}[c]{@{}l@{}}number of \\ classes\end{tabular} & \begin{tabular}[c]{@{}l@{}}Rotated \\ test images?\end{tabular} & \begin{tabular}[c]{@{}l@{}}training \\ images\end{tabular} & \begin{tabular}[c]{@{}l@{}}test \\ images\end{tabular} \\ \hline
outex TC10 & 24   & yes       & 480      & 3840    \\ \hline
outex TC11 & 24   & no        & 480      & 480     \\ \hline
outex TC20 & 68   & yes       & 1360     & 10880   \\ \hline
outex TC21 & 68   & no        & 1360     & 1360    \\ \hline
\end{tabular}
\end{center}
\end{table}

Initially we used the same combination of moments $M_{pq}$ and weights as proposed by \cite{Mukundan2014}.
%For reference, the results are shown in table \ref{}.
A range of moments and weights were randomly generated. The limits for the weights were set between 0.1 to 20.
Several choices of $M_{pq}$ were used, and the best results collected and stored.

For each combination of moments and weights, an independent classifier was trained using Random Forests, with 10 decision trees each, and a minimum sample split of 2.
The samples were split using a 10-fold cross validation.
These characteristics for the training were kept constant because the goal was to compare the feature extraction, and how easy or hard it would be to train this particular choice of parameters for the LTM method.

\subsection{Results}

Seven main experiments were carried out. 
A total of 728 classifiers were trained and tested.
Only the best results for each Outex dataset are shown in tables \ref{ResultsOutex10}, \ref{ResultsOutex11}, \ref{ResultsOutex20} and \ref{ResultsOutex21}.

%TODO table 
% 8 rows, 4 columns
\begin{table}[]
\begin{center}
\caption{Best results for Outex10}
\label{ResultsOutex10}
\begin{tabular}{|c|c|c|c|}
\hline
Experiment & $M_{pq}$ used & Weights & Best accuracy Outex10 \\ \hline
1          &  $M_{00} M_{11} M_{22} M_{33} M_{44}$             &  .1 1 1 1 20     & 0.93  $\pm$ 0.03              \\ \hline
2          &  $M_{00} M_{01} M_{10} M_{22} M_{44}$             &  .1 1 1 1 10     & 0.94  $\pm$ 0.04              \\ \hline
3          &  $M_{01} M_{10} M_{02} M_{20} M_{03}$             &  5 1 1 15 15     & 0.93  $\pm$ 0.07              \\ \hline
4          &  $M_{30} M_{33} M_{04} M_{40} M_{44}$             & 1 1 1 5 5        & 0.71  $\pm$ 0.10              \\ \hline
5          &  $M_{00} M_{01} M_{10} M_{11} M_{20}$             &  .1 5 5 5 5      & \textbf{0.96  $\pm$ 0.03}     \\ \hline
6          &  $M_{01} M_{10} M_{20} M_{12} M_{21}$             &  1 1 1 1 1       & 0.92  $\pm$ 0.07             \\ \hline
7          &  $M_{00} M_{22} M_{44} M_{22} M_{44}$             & .1 2 5 10 15     & 0.76  $\pm$ 0.05             \\ \hline
\end{tabular}
\end{center}
\end{table}

\begin{table}[]
\begin{center}
\caption{Best results for Outex11}
\label{ResultsOutex11}
\begin{tabular}{|c|c|c|c|}
\hline
Experiment & $M_{pq}$ used & Weights & Best accuracy Outex11 \\ \hline
1          &  $M_{00} M_{11} M_{22} M_{33} M_{44}$             &  .1 1 1 1 20     &  0.91  $\pm$ 0.02             \\ \hline
2          &  $M_{00} M_{01} M_{10} M_{22} M_{44}$             & .1 1 1 10 1      &  0.96  $\pm$ 0.02              \\ \hline
3          &  $M_{01} M_{10} M_{02} M_{20} M_{03}$             &  1 1 1 5 5       &  0.97  $\pm$ 0.02             \\ \hline
4          &  $M_{30} M_{33} M_{04} M_{40} M_{44}$             &  1 1 1 1 1       &  0.84  $\pm$ 0.04            \\ \hline
5          &  $M_{00} M_{01} M_{10} M_{11} M_{20}$             & .1 1 1 1 20      &  0.97  $\pm$ 0.02              \\ \hline
6          &  $M_{01} M_{10} M_{20} M_{12} M_{21}$             & 1 1 1 1 1        & \textbf{0.97  $\pm$ 0.02}              \\ \hline
7          &  $M_{00} M_{22} M_{44} M_{22} M_{44}$             &  .1 2 5 10 15    &  0.78  $\pm$ 0.05             \\ \hline
\end{tabular}
\end{center}
\end{table}

\begin{table}[]
\begin{center}
\caption{Best results for Outex20}
\label{ResultsOutex20}
\begin{tabular}{|c|c|c|c|}
\hline
Experiment & $M_{pq}$ used & Weights & Best accuracy Outex20 \\ \hline
1          &  $M_{00} M_{11} M_{22} M_{33} M_{44}$             &   .1 1 1 1 20      & 0.80  $\pm$ 0.04              \\ \hline
2          &  $M_{00} M_{01} M_{10} M_{22} M_{44}$             & .1 1 1 10 20       & \textbf{0.83  $\pm$ 0.04}              \\ \hline
3          &  $M_{01} M_{10} M_{02} M_{20} M_{03}$             & 1 1 1 5 5          & 0.71  $\pm$ 0.06              \\ \hline
4          &  $M_{30} M_{33} M_{04} M_{40} M_{44}$             &  1 1 1 1 1         & 0.43  $\pm$ 0.04            \\ \hline
5          &  $M_{00} M_{01} M_{10} M_{11} M_{20}$             &.1 1 1 1 5          & 0.82  $\pm$ 0.06             \\ \hline
6          &  $M_{01} M_{10} M_{20} M_{12} M_{21}$             &  1 1 1 1 1         & 0.69  $\pm$ 0.08              \\ \hline
7          &  $M_{00} M_{22} M_{44} M_{22} M_{44}$             & .1 2 5 10 15       & 0.59  $\pm$ 0.03            \\ \hline
\end{tabular}
\end{center}
\end{table}

\begin{table}[]
\begin{center}
\caption{Best results for Outex21}
\label{ResultsOutex21}
\begin{tabular}{|c|c|c|c|}
\hline
Experiment & $M_{pq}$ used & Weights & Best accuracy Outex21 \\ \hline
1          &  $M_{00} M_{11} M_{22} M_{33} M_{44}$             &  .1 1 1 1 20      & 0.78  $\pm$ 0.03              \\ \hline
2          &  $M_{00} M_{01} M_{10} M_{22} M_{44}$             &  .1 1 1 10 20     & \textbf{0.80  $\pm$ 0.02}   \\ \hline
3          &  $M_{01} M_{10} M_{02} M_{20} M_{03}$             &  1 1 1 5 5        & 0.73  $\pm$ 0.03             \\ \hline
4          &  $M_{30} M_{33} M_{04} M_{40} M_{44}$             &  1 20 1 1 1       & 0.48  $\pm$ 0.02             \\ \hline
5          &  $M_{00} M_{01} M_{10} M_{11} M_{20}$             & .1 2 5 10 15      & 0.79  $\pm$ 0.03              \\ \hline
6          &  $M_{01} M_{10} M_{20} M_{12} M_{21}$             &  1 1 20 1 1       & 0.74  $\pm$ 0.03             \\ \hline
7          &  $M_{00} M_{22} M_{44} M_{22} M_{44}$             & .1 2 5 10 15      & 0.56  $\pm$ 0.05              \\ \hline
\end{tabular}
\end{center}
\end{table}

The results show that using both low and high order moments get the best results for larger number of classes (Outex 20 and 21).
However, for a smaller number of classes (Outex 10 and 11) the best combination of $M_{pq}$ was using low moment orders.
In terms of the weights, low order moments seem to represent the texture better if using low weights ($< 1$), while higher order benefit from having their weights increased.
For example, one of the best results for Outex 20 and 21 used weights 10 and 20 for the moments $M_{22}$ and $M_{44}$ respectively (see tables \ref{ResultsOutex20} and \ref{ResultsOutex21}).

%After varying the moments and the weights, we collected the best results for those combinations.
%These can be seen in figure \ref{}.
The comparison of the LTM results with those of the various LBP methods is shown in table \ref{ResultsLBP}.

\begin{table}[]
\begin{center}
\caption{Comparative results for LTM and LBP methods.}
\label{ResultsLBP}
\begin{tabular}{|l|c|c|c|c|}
\hline
                                & Outex10                   & Outex11                    & Outex20                    & Outex21 \\ \hline
LTM~\cite{Mukundan2014}         & \textbf{0.96  $\pm$ 0.03 }& 0.97  $\pm$ 0.02           & \textbf{0.83  $\pm$ 0.04}  & \textbf{0.80  $\pm$ 0.02}  \\ \hline
OLBP~\cite{OjalaEtAl2002}       & 0.94  $\pm$ 0.06          & \textbf{0.97  $\pm$ 0.01}  & 0.67  $\pm$ 0.09           & 0.78  $\pm$ 0.01  \\ \hline
CS-LBP~\cite{HeikkilaEtAl2006}  & 0.87  $\pm$ 0.06          & 0.93  $\pm$ 0.04           & 0.67  $\pm$ 0.06           & 0.67  $\pm$ 0.04  \\ \hline
CS-LDP ~\cite{XueEtAl2011}      & 0.85  $\pm$ 0.07          & 0.91  $\pm$ 0.04           & 0.64  $\pm$ 0.06           & 0.65  $\pm$ 0.04  \\ \hline
XCS-LBP~\cite{SilvaEtAl2015}    & 0.85  $\pm$ 0.05          & 0.82  $\pm$ 0.10           & 0.54  $\pm$ 0.04           & 0.49  $\pm$ 0.05  \\ \hline
\end{tabular}
\end{center}
\end{table}

The Random Forest classifiers were created with a relatively low number of decision trees.%, so the classification results may not be the best compared to other published results.
Despite that fact, the LTM method got reasonably good results, and apart from Outex11 (for which is almost a draw with OLBP), it got the best classification results compared to the LBP based methods.

%\section{LTM modifications}
%In the original article, Mukundan used a 5x5 kernel, which implies in having moments to the $8^{th}$ order.

%TODO This approach can then be compared to LBPs....

\section{Experimenting with Different Kernel Sizes}
In the original LTM paper only the kernel size 5x5 was used.
We decide to experiment with different sizes, 3x3 and 7x7.

When using a different kernel size, the order of the moments $M_{pq}$ are limited to a certain combination.
%Table \ref{Orders} shows the possible orders for a 7x7 kernel. Both the 3x3 and 5x5 are contained within the 7x7 list.
The table shows that in 3x3 kernels only 9 orders can be used, while for 5x5 kernels there are 25 orders and for 7x7 kernels there are 49 orders.
The maximum order for 3x3 kernels is $M_{22}$, for 5x5 kernels is $M_{44}$ and for 7x7 kernels it is $M_{66}$.
Table \ref{Orders} shows the possible orders for 3x3, 5x5 and 7x7 kernels.

We repeated the same training process used in the previous section.
A total of 1456 classifiers using Random Forests was produced, 728 for the 3x3 kernel size, and another 728 for the 7x7 kernel size.

\begin{table}[]
\caption{A complete list of different orders for different kernel sizes. In light grey there are all the possible moment orders for a 3x3 kernel (9 moments).
In dark gray the additional $pq$ are shown for 5x5 kernels (25 moments), and in black the additional orders for a 7x7 kernel (49 moments) are shown.}
\label{Orders}
\centering
\resizebox{\textwidth}{!}{%
\begin{tabular}{l|lllllllllllll}
\hline
Order (p+q) & 0  & 1  & 2  & 3  & 4  & 5  & 6  & 7  & 8  & 9  & 10 & 11 & 12 \\ \hline
            & \color{lightgray}{00} & \color{lightgray}{01} & \color{lightgray}{11} & \color{lightgray}{12} & \color{lightgray}{22} & \color{gray}{23} & \color{gray}{33} & \color{gray}{34} & \color{gray}{44} & 45 & 55 & 56 & 66 \\
            &    & \color{lightgray}{10} & \color{lightgray}{02} & \color{lightgray}{21} & \color{gray}{13} & \color{gray}{32} & \color{gray}{24} & \color{gray}{43} & 35 & 54 & 46 & 65 &    \\
            &    &    & \color{lightgray}{20} & \color{gray}{03} & \color{gray}{31} & \color{gray}{14} & \color{gray}{42} & 25 & 53 & 36 & 64 &    &    \\
            &    &    &    & \color{gray}{30} & \color{gray}{04} & \color{gray}{41} & 15 & 52 & 26 & 63 &    &    &    \\
            &    &    &    &    & \color{gray}{40} & 05 & 51 & 16 & 62 &    &    &    &    \\
            &    &    &    &    &    & 50 & 06 & 61 &    &    &    &    &    \\
            &    &    &    &    &    &    & 60 &    &    &    &    &    &    \\ \hline
\end{tabular}%
}
\end{table}

The best results for 3x3 kernels is presented in tables \ref{ResultsOutex10_3x3}, \ref{ResultsOutex11_3x3}, \ref{ResultsOutex20_3x3} and \ref{ResultsOutex21_3x3}.
The best results for 7x7 kernels is presented in tables \ref{ResultsOutex10_7x7}, \ref{ResultsOutex11_7x7}, \ref{ResultsOutex20_7x7} and \ref{ResultsOutex21_7x7}.

\begin{table}[]
\begin{center}
\caption{Best results for Outex10 using 3x3 kernels.}
\label{ResultsOutex10_3x3}
\begin{tabular}{|c|c|c|c|}
\hline
Experiment & $M_{pq}$ used & Weights & Best accuracy Outex10 \\ \hline
1          &  $M_{00} M_{01} M_{10} M_{21} M_{22}$             & .1 1 1 1 20      &  0.91  $\pm$ 0.08   \\ \hline
2          &  $M_{00} M_{01} M_{10} M_{20} M_{22}$             & .1 1 1 10 20     &  \textbf{0.95  $\pm$ 0.03}   \\ \hline
3          &  $M_{01} M_{10} M_{02} M_{20} M_{22}$             & 1 1 1 5 5        &  0.93  $\pm$ 0.06   \\ \hline
4          &  $M_{00} M_{11} M_{02} M_{20} M_{22}$             & .1 2 5 10 15     &  0.92  $\pm$ 0.05    \\ \hline
5          &  $M_{00} M_{01} M_{10} M_{11} M_{20}$             & .1 1 1 1 10      &  0.95  $\pm$ 0.04    \\ \hline
6          &  $M_{01} M_{10} M_{20} M_{12} M_{21}$             &  1 1 1 5 5       &  0.91  $\pm$ 0.07   \\ \hline
7          &  $M_{00} M_{11} M_{22} M_{11} M_{22}$             & .1 2 5 10 15     &  0.83  $\pm$ 0.04   \\ \hline
\end{tabular}
\end{center}
\end{table}

\begin{table}[]
\begin{center}
\caption{Best results for Outex11 using 3x3 kernels.}
\label{ResultsOutex11_3x3}
\begin{tabular}{|c|c|c|c|}
\hline
Experiment & $M_{pq}$ used & Weights & Best accuracy Outex11 \\ \hline
1          &  $M_{00} M_{11} M_{20} M_{21} M_{22}$             & .1 2 5 10 15 & 0.96  $\pm$ 0.02 \\ \hline
2          &  $M_{00} M_{01} M_{10} M_{20} M_{22}$             & .1 1 1 10 5  & 0.96  $\pm$ 0.01 \\ \hline
3          &  $M_{01} M_{10} M_{02} M_{20} M_{22}$             & 1 10 10 10 5 & 0.97  $\pm$ 0.03 \\ \hline
4          &  $M_{00} M_{11} M_{02} M_{20} M_{22}$             & .1 1 1 10 5  & 0.95  $\pm$ 0.04\\ \hline
5          &  $M_{00} M_{01} M_{10} M_{11} M_{20}$             & .1 5 5 5 5   & \textbf{0.97  $\pm$ 0.02} \\ \hline
6          &  $M_{01} M_{10} M_{20} M_{12} M_{21}$             & 1 1 1 5 5    & 0.96  $\pm$ 0.01\\ \hline
7          &  $M_{00} M_{11} M_{22} M_{11} M_{22}$             & .1 2 5 10 15 & 0.88  $\pm$ 0.06\\ \hline
\end{tabular}
\end{center}
\end{table}

\begin{table}[]
\begin{center}
\caption{Best results for Outex20 using 3x3 kernels.}
\label{ResultsOutex20_3x3}
\begin{tabular}{|c|c|c|c|}
\hline
Experiment & $M_{pq}$ used & Weights & Best accuracy Outex20 \\ \hline
1          &  $M_{00} M_{01} M_{10} M_{21} M_{22}$             & .1 1 1 1 20  & 0.81  $\pm$ 0.07 \\ \hline 
2          &  $M_{00} M_{01} M_{10} M_{20} M_{22}$             & .1 2 5 10 15 & \textbf{0.84  $\pm$ 0.03} \\ \hline
3          &  $M_{01} M_{10} M_{02} M_{20} M_{22}$             & .1 1 1 1 10  & 0.73  $\pm$ 0.05 \\ \hline 
4          &  $M_{00} M_{11} M_{02} M_{20} M_{22}$             & .1 2 5 10 15 & 0.80  $\pm$ 0.04\\ \hline
5          &  $M_{00} M_{01} M_{10} M_{11} M_{20}$             &.1 2 5 10 15  & 0.82  $\pm$ 0.05\\ \hline 
6          &  $M_{01} M_{10} M_{20} M_{12} M_{21}$             & 1 1 1 5 5    & 0.70  $\pm$ 0.08\\ \hline   
7          &  $M_{00} M_{11} M_{22} M_{11} M_{22}$             & .1 2 5 10 15 & 0.74  $\pm$ 0.04\\ \hline
\end{tabular}
\end{center}
\end{table}

\begin{table}[]
\begin{center}
\caption{Best results for Outex21 using 3x3 kernels.}
\label{ResultsOutex21_3x3}
\begin{tabular}{|c|c|c|c|}
\hline
Experiment & $M_{pq}$ used & Weights & Best accuracy Outex21 \\ \hline
1          &  $M_{00} M_{01} M_{10} M_{21} M_{22}$             & .1 1 1 1 20  & 0.79  $\pm$ 0.04 \\ \hline 
2          &  $M_{00} M_{01} M_{10} M_{20} M_{22}$             & .1 1 1 10 10 & \textbf{0.81  $\pm$ 0.03} \\ \hline
3          &  $M_{01} M_{10} M_{02} M_{20} M_{22}$             & 1 1 1 5 5    & 0.75  $\pm$ 0.03\\ \hline   
4          &  $M_{00} M_{11} M_{02} M_{20} M_{22}$             & .1 1 1 10 20 & 0.77  $\pm$ 0.04 \\ \hline
5          &  $M_{00} M_{01} M_{10} M_{11} M_{20}$             & .1 2 5 10 15 & 0.79  $\pm$ 0.04\\ \hline
6          &  $M_{01} M_{10} M_{20} M_{12} M_{21}$             & 1 1 1 1 1    & 0.74  $\pm$ 0.03 \\ \hline   
7          &  $M_{00} M_{11} M_{22} M_{11} M_{22}$             & .1 2 5 10 15 & 0.69  $\pm$ 0.06\\ \hline
\end{tabular}
\end{center}
\end{table}

\begin{table}[]
\begin{center}
\caption{Best results for Outex10 using 7x7 kernels.}
\label{ResultsOutex10_7x7}
\begin{tabular}{|c|c|c|c|}
\hline
Experiment & $M_{pq}$ used & Weights & Best accuracy Outex10 \\ \hline
1          &  $M_{00} M_{11} M_{22} M_{33} M_{44}$             &  .1 1 1 1 20    & 0.93  $\pm$ 0.04  \\ \hline 
2          &  $M_{00} M_{01} M_{10} M_{22} M_{44}$             &  .1 1 1 1 20    & 0.94  $\pm$ 0.04  \\ \hline 
3          &  $M_{01} M_{10} M_{02} M_{20} M_{03}$             &   1 1 1 1 1     & 0.93  $\pm$ 0.05  \\ \hline 
4          &  $M_{30} M_{33} M_{04} M_{40} M_{44}$             &   .1 1 1 1 1    & 0.79  $\pm$ 0.10  \\ \hline 
5          &  $M_{00} M_{01} M_{10} M_{11} M_{20}$             &   .1 1 1 1 1    & \textbf{0.96  $\pm$ 0.03}  \\ \hline 
6          &  $M_{01} M_{10} M_{20} M_{12} M_{21}$             &  1 1 1 1 1      & 0.93  $\pm$ 0.06  \\ \hline 
7          &  $M_{00} M_{22} M_{44} M_{22} M_{44}$             &  .1 2 5 10 15   & 0.77  $\pm$ 0.06  \\ \hline 
\end{tabular}
\end{center}
\end{table}

\begin{table}[]
\begin{center}
\caption{Best results for Outex11 using 7x7 kernels.}
\label{ResultsOutex11_7x7}
\begin{tabular}{|c|c|c|c|}
\hline
Experiment & $M_{pq}$ used & Weights & Best accuracy Outex11 \\ \hline
1          &  $M_{00} M_{01} M_{10} M_{22} M_{44}$             & .1 1 1 1 20  & 0.92  $\pm$ 0.02 \\ \hline 
2          &  $M_{00} M_{01} M_{10} M_{22} M_{44}$             & .1 1 1 10 5  & 0.94  $\pm$ 0.02 \\ \hline 
3          &  $M_{01} M_{10} M_{02} M_{20} M_{03}$             & 1 1 1 1 1    & 0.95  $\pm$ 0.02 \\ \hline 
4          &  $M_{30} M_{33} M_{04} M_{40} M_{44}$             & .5 5 1 1 1   & 0.90  $\pm$ 0.04 \\ \hline 
5          &  $M_{00} M_{01} M_{10} M_{11} M_{20}$             & .1 1 1 1 2   & \textbf{0.97  $\pm$ 0.02} \\ \hline 
6          &  $M_{01} M_{10} M_{20} M_{12} M_{21}$             & 1 1 1 1 1    & \textbf{0.97  $\pm$ 0.02} \\ \hline
7          &  $M_{00} M_{22} M_{44} M_{22} M_{44}$             & .1 1 1 10 10 & 0.79  $\pm$ 0.04 \\ \hline 
\end{tabular}
\end{center}
\end{table}

\begin{table}[]
\begin{center}
\caption{Best results for Outex20 using 7x7 kernels.}
\label{ResultsOutex20_7x7}
\begin{tabular}{|c|c|c|c|}
\hline
Experiment & $M_{pq}$ used & Weights & Best accuracy Outex20 \\ \hline
1          &  $M_{00} M_{01} M_{10} M_{22} M_{44}$             & .1 1 1 1 20  & 0.80  $\pm$ 0.05 \\ \hline 
2          &  $M_{00} M_{01} M_{10} M_{22} M_{44}$             & .1 1 1 10 10 & 0.82  $\pm$ 0.04 \\ \hline 
3          &  $M_{01} M_{10} M_{02} M_{20} M_{03}$             & 1 1 1 5 5    & 0.67  $\pm$ 0.07 \\ \hline 
4          &  $M_{30} M_{33} M_{04} M_{40} M_{44}$             & 5 1 1 15 15  & 0.51  $\pm$ 0.06 \\ \hline 
5          &  $M_{00} M_{01} M_{10} M_{11} M_{20}$             & .1 1 1 10 1  & \textbf{0.80  $\pm$ 0.04} \\ \hline 
6          &  $M_{01} M_{10} M_{20} M_{12} M_{21}$             & 1 1 20 1 1   & 0.67  $\pm$ 0.07 \\ \hline 
7          &  $M_{00} M_{22} M_{44} M_{22} M_{44}$             & .1 2 5 10 15 & 0.64  $\pm$ 0.03 \\ \hline 
\end{tabular}
\end{center}
\end{table}

\begin{table}[]
\begin{center}
\caption{Best results for Outex21 using 7x7 kernels.}
\label{ResultsOutex21_7x7}
\begin{tabular}{|c|c|c|c|}
\hline
Experiment & $M_{pq}$ used & Weights & Best accuracy Outex21 \\ \hline
1          &  $M_{00} M_{01} M_{10} M_{22} M_{44}$             & .1 1 1 1 20  &  0.77  $\pm$ 0.04 \\ \hline
2          &  $M_{00} M_{01} M_{10} M_{22} M_{44}$             & .1 1 1 10 20 &  \textbf{0.78  $\pm$ 0.04} \\ \hline
3          &  $M_{01} M_{10} M_{02} M_{20} M_{03}$             & 1 1 1 5 5    &  0.71  $\pm$ 0.03\\ \hline 
4          &  $M_{30} M_{33} M_{04} M_{40} M_{44}$             & 1 1 1 5 5    &  0.59  $\pm$ 0.05\\ \hline 
5          &  $M_{00} M_{01} M_{10} M_{11} M_{20}$             & .1 2 5 10 15 &  0.77  $\pm$ 0.03\\ \hline 
6          &  $M_{01} M_{10} M_{20} M_{12} M_{21}$             & 1 10 10 10 5 &  0.68  $\pm$ 0.03\\ \hline 
7          &  $M_{00} M_{22} M_{44} M_{22} M_{44}$             & .1 2 5 10 15 &  0.57  $\pm$ 0.05\\ \hline 
\end{tabular}
\end{center}
\end{table}

The same general trend for the weights was observed for these kernel sizes.
The best results for each dataset is marked in each table.

We also tabulated the best results for each kernel size (table \ref{ResultsPerKernel}).
Unexpectedly the results are slightly better for the smaller 3x3 kernel.
The difference is very small though and is within the error margin, so statistically there is no difference for different kernels.

\begin{table}[]
\begin{center}
\caption{Comparing the best results for each kernel.}
\label{ResultsPerKernel}
\begin{tabular}{|c|c|c|c|}
\hline
                & 3x3         & 5x5  &    7x7   \\ \hline
Outex 10        & 0.95  $\pm$ 0.03           & \textbf{0.96  $\pm$ 0.03}     & \textbf{0.96  $\pm$ 0.03}            \\ \hline
Outex 11        & \textbf{0.97  $\pm$ 0.02}            & \textbf{0.97  $\pm$ 0.02}     & \textbf{0.97  $\pm$ 0.02}             \\ \hline
Outex 20        & \textbf{0.84  $\pm$ 0.03}          & 0.83  $\pm$ 0.04     & 0.80  $\pm$ 0.04           \\ \hline 
Outex 21        & \textbf{0.81  $\pm$ 0.03}          & 0.80  $\pm$ 0.02     & 0.78  $\pm$ 0.04           \\ \hline 
\end{tabular}
\end{center}
\end{table}

\section{Conclusions and Future work}

In this paper we used machine learning to train classifiers to assess the LTM method as a feature extractor for texture classification.
We were able to conclude that the weights for each local moment should be higher for high orders and lower for low orders.
There was no difference in the accuracy of the classifier when changing the kernel size.

To keep the experiments within the limited time frame we had, we did not yet test higher order moments when using the 7x7 kernel size.
It is possible that this makes some difference, and only additional experiments will confirm that. 
% \subsection{Extending the Histogram}
% 
We would like to extend the histogram in order to use more than 5 moments to compose the feature space.
There is an issue with the number of permutations generated depending on the number of moments used.
Using 5 moments, the total number of permutations is $5!=120$, which generates pixel values for the LTM image from 0 to 119.
Using 6 moments would require $6!=720$ pixel values, 7 moments would require $7!=5040$ and so on.
Obviously it would be impossible to use too many moments in this manner, as for example 25 moments would have $25!=1.55 \times 10^{25}$ permutations.
It should be possible to extend the feature space, using up to 8 or 9 moments, using the histogram directly without generating an LTM image.
% So we decided to settle with a maximum of 8 moments ($8!=40320$) as a feasible number.
% The idea is to compare the classification and measure whether more moments improves the accuracy of the texture classification.
% 
% Unfortunately, this is not the only choice to make. 
% There is also the question of which specific moments one can choose.
% For the extended histogram experiments, we also assessed the difference in accuracy when using only the symmetric kernels (where $p=q$ for $T_{pq}$) and using a mixture of asymmetric and symmetric kernels.

\bibliographystyle{IEEEtran}
\bibliography{LTM}
\end{document}